\documentclass[pmlr]{jmlr}

\usepackage{longtable}

\usepackage{booktabs}
\usepackage{microtype}
\usepackage{graphicx}
\usepackage{caption}

\usepackage{float} 
\usepackage{multirow}

\usepackage{hyperref}
\usepackage{graphicx}
\usepackage{makecell}

\usepackage{amsmath}
\usepackage{amssymb}
\usepackage{mathtools}
\usepackage{bbm}

\usepackage[capitalize,noabbrev]{cleveref}
 
\usepackage[load-configurations=version-1]{siunitx} 

\usepackage[capitalize,noabbrev]{cleveref}

\newcommand{\indep}{\rotatebox[origin=c]{90}{$\models$}}
\newcommand{\E}{\mathbb{E}}

\makeatletter
\def\set@curr@file#1{\def\@curr@file{#1}} 
\makeatother

\theorembodyfont{\upshape}
\theoremheaderfont{\scshape}
\theorempostheader{:}
\theoremsep{\newline}

\jmlrvolume{182}
\jmlryear{2022}
\jmlrworkshop{Machine Learning for Healthcare}

\title[SurvLatent ODE : A Neural ODE based time-to-event model for longitudinal data]{SurvLatent ODE : A Neural ODE based time-to-event model with competing risks for longitudinal data improves cancer-associated Venous Thromboembolism (VTE) prediction}

\author{\Name{Intae Moon}
      \Email{itmoon@mit.edu}\\ 
       \addr Electrical Engineering and Computer Science\\
       Massachusetts Institute of Technology\\
       Cambridge, MA, USA
      \AND
      \Name{Stefan Groha}
      \Email{stefanm\textunderscore groha@dfci.harvard.edu}\\ 
      \addr Division of Population Sciences\\
      Dana-Farber Cancer Institute\\
      Boston, MA, USA
       \AND
      \Name{Alexander Gusev}
      \Email{alexander\textunderscore gusev@dfci.harvard.edu}\\ 
      \addr Division of Population Sciences\\
      Dana-Farber Cancer Institute\\
      Boston, MA, USA} 

\begin{document}

\maketitle

\begin{abstract}
  Effective learning from electronic health records (EHR) data for prediction of clinical outcomes is often challenging because of features recorded at irregular timesteps and loss to follow-up as well as competing events such as death or disease progression. To that end, we propose a generative time-to-event model, SurvLatent ODE, which adopts an Ordinary Differential Equation-based Recurrent Neural Networks (ODE-RNN) as an encoder to effectively parameterize dynamics of latent states under irregularly sampled input data. Our model then utilizes the resulting latent embedding to flexibly estimate survival times for multiple competing events without specifying shapes of event-specific hazard function. We demonstrate competitive performance of our model on MIMIC-III, a freely-available longitudinal dataset collected from critical care units, on predicting hospital mortality as well as the data from the Dana-Farber Cancer Institute (DFCI) on predicting onset of Venous Thromboembolism (VTE), a life-threatening complication for patients with cancer, with death as a competing event. SurvLatent ODE outperforms the current clinical standard Khorana Risk scores for stratifying VTE risk groups, while providing clinically meaningful and interpretable latent representations.
\end{abstract}


\section{Introduction} 
Electronic Health Records (EHR), which consist of longitudinal measurements of patients' measurements often extending over decades, offer unique opportunities to build data-driven clinical decision support tools. Accurate stratification of future disease can allow clinicians to focus clinical resources on the most at-risk patients while minimizing potential side effects and overtreatment for low-risk patients \citep{Myers2017, Maniruzzaman2018, Zeiberg2019}. However, most existing patient stratification tools still rely on simple, outcome-specific scoring systems that utilize a small number of static features \citep{Wilson1998, Khorana2008}. As patients are often treated over the course of multiple interactions, a data-driven clinical decision model should relate the heterogeneous temporal evolution of clinical measurements to the clinical outcome in a flexible form and provide updated outcome prediction as a patient's condition and clinical features change. As EHR data is typically collected incidentally, the model should appropriately handle data missingness due to irregular spacing of visits or loss to follow-up, as well as informative missingness – where the collection of a measurement may itself be predictive of a future outcome. When estimating risk of a clinical outcome, an effective clinical decision model should also consider potential competing events a patient may experience and update the risk for the outcome of interest accordingly. This enables clinicians to jointly consider the risk of cancer-specific death or a treatment-related toxicity when planning a therapeutic intervention. 

One widely utilized risk prediction framework is a time-to-event (or survival) analysis, which models the lifetime of a patient until a failure event. Importantly, time-to-event analysis can directly model outcomes that are lost to follow-up (also known as right-censoring), which is very common in EHR due to the lack of formal patient recruitment. One of the most widely utilized time-to-event models is the Cox proportional hazard model, a multivariable, semi-parametric framework proposed by \cite{Cox1972}, which relates log of relative hazard to a linear function of baseline features. Recently, there have been efforts to augment the Cox proportional hazard model by learning a non-linear relationship of baseline features to outcomes through deep neural networks \citep{Katzman2018, Chirag2021DeepCox}. However, these models make time-to-event predictions based on a single snapshot of features and fail to capture evolution of time-varying features, often informative of predicting disease onset. To address this gap, time-to-event methods capable of handling time-varying, longitudinal measurements have been proposed. For example, some models utilized class of Recurrent Neural Network (RNN) models to learn input temporal representation which then renders estimated time-to-event predictions \citep{Ren2018, Lee2020DynamicDeepHit, Chirag2021DeepParam}, while other models utilized a deep generative approach to model the event time distribution in a Bayesian framework \citep{Ranganath2016, Miscouridou2018}. However, presented with missing measurements, these models rely on strategies agnostic to latent time-dependent dynamics between observations and may lead to ill-defined latent representation.

In a multivariate time-series framework, many approaches have been developed to handle informative missingness to improve prediction performance \citep{Choi2016, Lipton2016, Futoma2017, Soleimani2017, Che2018Rnn}. One approach to deal with missing values involves deploying a decay mechanism on both input data and RNN hidden states for describing dynamics under missing observations \citep{Che2018Rnn}. Other methods utilized a Gaussian process (GP)-based framework to learn dynamics across observed measurements and provide continuous latent states \citep{Futoma2017, Soleimani2017}. However, these models rely on parametric assumptions on the latent dynamics, specified by either exponential decay functions or GP kernel functions with a stationary property and may not necessarily describe true latent dynamics. On the other hand, a recently proposed family of Neural Ordinary Differential Equations (Neural ODE) based models \citep{Chen2018, Rubanova2019} learns the latent dynamics using highly flexible neural network parameterized functions and is capable of handling input data which may arrive at arbitrary time points, providing an avenue to model longitudinal data with missing values without strong parametric assumptions on the latent dynamics.

In this work, we present a generative, time-to-event model, SurvLatent ODE, which utilizes the Neural ODE framework to effectively learn temporal dynamics of the input representation under irregularly sampled measurements. Adopting a multi-task learning framework, our proposed model allows the underlying mechanism of multiple events to be shared in the latent representation while utilizing a cause-specific decoder module to flexibly learn signals specific to each event from the shared latent representation. As a result, the proposed model provides survival functions for each patient across multiple clinical outcomes they may encounter while incorporating temporal evolution of their features. Using MIMIC-III, a publicly available longitudinal dataset collected at critical care units, we demonstrate that the proposed model significantly outperforms conventional as well as state-of-the-art time-to-event models in predicting time to hospital mortality. Furthermore, we utilize the longitudinal data from the Dana-Farber Cancer Institute (DFCI) for predicting time to VTE with all-cause mortality as a competing event, and show that the proposed model significantly outperforms the current clinical standard, Khorana scores \citep{Khorana2008}. Finally, we demonstrate the learned latent representation offers interpretable clusters of patients with meaningfully different outcomes. The implementation of SurvLatent ODE is available at \href{https://github.com/itmoon7/survlatent_ode}{\url{https://github.com/itmoon7/survlatent_ode}}.

\subsection*{Generalizable Insights about Machine Learning in the Context of Healthcare}

\begin{itemize}
  \item To the best of our knowledge, our proposed model is the first demonstration of the ODE-based variational autoencoder time-to-event model for longitudinal data where temporal latent dynamics in the input data are explicitly modeled via neural networks. 
  \item Combined with the effective longitudinal modeling, our proposed framework enables flexible estimation of hazard functions for the event of interest as well as competing events via a multi-task learning framework, which gains significant improvements over conventional and recently published deep learning based survival models.
  \item Applied to the in-house dataset of Venous Thromboembolism (VTE) events, our model significantly outperforms current clinical standards, Khorana scores \citep{Khorana2008} and provides insights into influential features for elevated VTE risks via interpretable latent representations. 
\end{itemize}

\section{Related Work}

Deep learning based approaches have been widely utilized in many published work in the context of time-to-event analysis to learn complex, non-linear relationship between features via neural networks \citep{Lee2018DeepHit, Katzman2018, Chirag2021DeepCox}, Convolutional Neural Network (CNN) \citep{Jarrett2018}, Gaussian Process \citep{Alaa2017}, and Recurrent Neural Networks \citep{Ren2018}. However, all of the above models only consider patients' time static features at baseline. More recently, several methods have been developed to incorporate time-varying features. Recurrent Deep Survival Machines \citep{Chirag2021DeepParam} incorporates time-varying features from longitudinal data and estimates conditional survival distribution through a fixed mixture of parametric distributions like Weibull or Log-Normal. However, a parametric assumption on the underlying time-to-event process may lead to a model miss-specification and limit the flexibility of relating neural network-learned representations to the conditional survival estimates. Furthermore, the model only handles a single risk, which may lead to an overestimation of the disease risk since in healthcare applications there is often a varying degree of dependence across competing events \citep{Berry2010, Austin2016}. \cite{Lee2020DynamicDeepHit} proposed Dynamic-Deephit, a discrete time-to-event model, which incorporates time-varying features and utilizes a multi-task learning framework to estimate joint distribution of the first hitting time and competing events. A key assumption of the model is that a patient experiences an event over the predefined finite time horizon with a probability of 1, which often does not hold in real data. When handling missing values, both models rely on population-level statistics (e.g. means) and/or missing data indicators \citep{Lipton2016}, which does not consider dynamics of underlying patient-specific health trajectory. \cite{Ranganath2016} and \cite{Miscouridou2018} demonstrated time-to-event models where missing observations were handled by a shared latent process which models observed measurements as well as event times in a Bayesian framework. However, their models relied on strong exponential parametric assumption on the underlying data generating process and were not evaluated in the longitudinal setting. 

In the context of more conventional time-to-event frameworks for competing risks, cause-specific Cox regression and Fine-Gray Cox regression \citep{Fine1999} models are widely used for the analysis of continuous event times. These two models have been extended to the analysis involving event times measured on a discrete time scale \citep{Lee2018, Berger2020}. While the Fine-Gray model enables estimation of covariates' effects on the cumulative incidence function for the event of interest, the cause-specific Cox model captures effect of covariates on the cause-specific hazard function for each event, which denotes the instantaneous rate of the corresponding event occurrence for patients who are currently event free \citep{Putter2007, Austin2016}. Both models assume linear relationship between patients' covariates and log of the relative hazard as well as proportional hazard where effects of covariates on the relative hazards remain constant over time. However, these assumptions are limiting because true underlying time-to-event processes are often described by complex nonlinear relationships between patients' biomarkers and effects of biomarkers on the survival may change over time as patients' health status are changing. 

Ordinary Differential equations parameterized by neural networks, popularized by \cite{Chen2018}, have been utilized in time-to-event analysis in a multi-state setting \citep{Groha2021} and a single-event continuous-time setting \citep{Tang2020}. \cite{Groha2021} proposes a Neural ODE approach to estimate the Kolmogorov forward equations which then provides transition probabilities in a multi-state framework. \cite{Tang2020} models the distribution of survival time for a single event by learning the dynamics of the cumulative hazard function via neural networks. However, both approaches only considered the baseline data and left out informative signals from time-varying features. To the best of our knowledge, our proposed model is the first demonstration of full ODE-based encoder-decoder architecture for modeling longitudinal data in a time-to-event framework.

\section{Methods}
\subsection{Preliminaries}

We first introduce three key modeling choices of the proposed framework : Neural Ordinary Differential Equations (Neural ODEs), discrete time-to-event analysis, and competing risks. 

\subsubsection{Neural Ordinary Differential Equations}
First, we model a patient's health trajectory using Neural Ordinary Differential Equations (Neural ODEs). Neural ODEs are a recently proposed family of neural networks-based continuous time models \citep{Chen2018}, which parameterize the dynamics of a hidden state $h(t)$. This hidden state may represent the latent health trajectory of a patient and a function of their time-varying features (i.e. $h(t) = z_1(X^t)$, where $z_1(\cdot)$ may be neural networks and $X^t$ is a set of patient's features up until time $t$). Given that the function $f_{\theta}$ (parameterized by the neural network) specifies the dynamics of the hidden state, we can define the hidden state $h(t)$ (e.g. a latent health trajectory) at any arbitrary time $t$ as follows:
\begin{align}
    \label{eqn:node_}
    h(t) = h(t_0) + \int_{t_0}^{t} f_{\theta}(h(\tau), \tau) d\tau,
\end{align} where $h(t_0)$ is an initial hidden state. \cite{Rubanova2019} demonstrated that this Neural ODE framework can be incorporated into a Recurrent Neural Network (RNN) model, allowing $f_{\theta}$ to learn the dynamics of the RNN hidden state $h(t)$ such that $h(t)$ is well-defined in a continuous time. Such an ODE-RNN model can incorporate irregularly sampled input data without having to explicitly impute missing values. Finally, the hidden state trajectory may be further decoded into, for example, the hazard function trajectory for a time-to-event prediction.

\subsubsection{Discrete time-to-event analysis}
Second, we model the outcome event times as a discrete time process. A common constraint of healthcare data is that event times such as time of hospital death and time of stroke diagnosis are often recorded on a discrete time scale, for example, in hours or days. In this case, the exact time of an event is not known and all we know is that the event occurred between two consecutive time points. Formally, a continuous time horizon is divided into discrete intervals (i.e. $[0, t_1), [t_1, t_2), ...$), where the time resolution is often determined by a domain knowledge, and an event time is denoted by $T$, where $T = t$ means that the event of interest has happened in the interval $[t-1, t)$ \citep{Tutz2016}.

\subsubsection{Competing risks}

Third, we seek to model multiple competing outcomes (e.g. cancer-specific death, metastasis, and VTE) that a patient is under risk for simultaneously. One common approach to handle competing risks in a discrete time-to-event analysis is to model the cause-specific discrete hazard function (i.e. rate of the event occurrence in a current time interval for event free patients in the previous time interval) \citep{Austin2016}. At a patient level, this framework provides insights on the individual event mechanism by modeling the contribution of a patient's features on the hazard function over time. Formally, given a set of $b$ different events $\mathcal{K} = \{1, ..., b\}$ and a set of features $X$, the cause-specific discrete hazard function $\lambda_k(t)$ for event $k$ is
\begin{equation}
    \lambda_k(t|X) = P(T = t, K = k | T \geq t, X),
\end{equation} where $k \in \mathcal{K}$. And the resulting overall discrete hazard function is
\begin{equation}
    \lambda(t|X) = \sum_{k=1}^b \lambda_k(t|X) = P(T = t | T \geq t, X).
\end{equation} An overall event-free survival probability $S(t|X)$ is 
\begin{equation}
    \begin{split}
        S(t|X) &= P(T > t | X) = \prod_{\tau \leq t} (1 - \lambda(\tau|X)).
    \end{split}
\end{equation} Finally, the cause-specific cumulative incidence function (CIF) for event $k$, $F_k(t|X)$, which captures the risk for event $k$ at time $t$ under the other competing risks conditioned on $X$, is
\begin{equation}
    \begin{split}
        F_k(t|X) &= P(T \leq t, K = k|X) = \sum_\tau^t P(T = \tau, K = k|X) = \sum_\tau^t \lambda_k(\tau|X)S(\tau - 1|X).
    \end{split}
\end{equation}

\subsection{Notation and setting}
\label{ssec:notation}

In brief, our model takes as input a set of individuals with (possibly irregularly spaced) time-varying features and event times of one or more outcome events, and learns the functions that map the features to each event specific hazard over time. We formalize these variables below.

Survival dataset $\mathcal{D}$ with sample size $N$ is a set of tuples $\{ (t_i, k_i, \delta_i, \mathcal{X}_i)\}^N_{i = 1}$ where $t_i$ and $k_i$ are the observed survival time and the event type for sample $i$, respectively. $\delta_i$ indicates whether the event occurred for sample $i$. Note that $t_i = \text{min}(\tilde{T}_i, C_i)$ where $\tilde{T}_i$ is the true event time and $C_i$ is the right censoring (i.e. loss to follow-up) time of sample $i$. Therefore, for right-censored samples (i.e. those with $\delta_i = 0$), $t_i$ corresponds to $C_i$. Note that throughout this paper, we assume that the censoring mechanism is non-informative. 

In the competing risks scenario, there may be multiple events where an occurrence of one of those events prevents the occurrence of the other events (for example, illness cannot occur after death). To that end, given the finite set of $b$ mutually exclusive competing events, $\mathcal{K} = \{1,...,b\}$, the event type of sample $i$ is denoted by $k_i$, where $k_i \in \mathcal{K}$. 

$\mathcal{X}_i$ is the set of features for sample $i$ longitudinally measured at some irregular timestamps denoted by $\tau_i$. In other words, given that $\tau_{i,0}$ and $\tau_{i,l}$ are the initial measurement time and latest measurement time for sample $i$, respectively, $\mathcal{X}_i = \{x_{i}(\tau_{i,0}), x_{i}(\tau_{i,1}), ..., x_{i}(\tau_{i,l})\}$, where $x_{i}(\tau_i) \in \mathbb{R}^M$ and $M$ is the number of features. Let $T^{r}$ be a random variable for the remaining time-to-event from the latest measurement or pre-defined landmark such as treatment start date. Our goal is to estimate the discrete cause-specific hazard function denoted by $\lambda_{k}(t | \mathcal{X}_i, Z_i^t) = P(T^{r} = t, K = k | T^{r} \geq t, \mathcal{X}_i, Z_i^t)$, where $Z_i^t$ is the latent trajectory representing the data $\mathcal{X}_i$ up until time $t$. For the rest of our paper, we use $\lambda_{i,k}^*(t)$ to denote $\lambda_{k}(t | \mathcal{X}_i, Z_i^t)$. Finally, we can relate $\lambda_{i,k}^*(t)$ to the overall event-free survival function, $S(t|\mathcal{X}_i)$, as well as cause-specific CIF, $F_k(t|\mathcal{X}_i)$, to assess individual risk of experiencing event $k$.

\subsection{SurvLatent ODE}
\label{sec:ode_rnn_Cox}
\subsubsection{Overview of the model architecture}
\label{ssec:surv_func_est}

\begin{figure*}[!ht]
  \centering
  \includegraphics[width=1\textwidth]{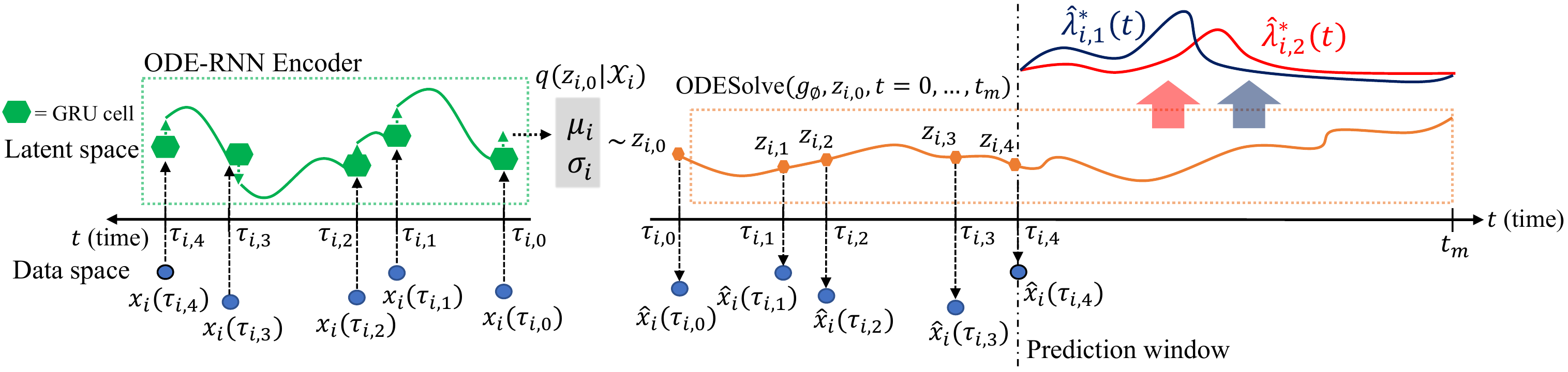}
  
  \caption{\label{fig:model_architecture_simple} \small \textit{Simplified} illustration of SurvLatent ODE taking irregularly sampled features of patient $i$, $\mathcal{X}_i$, and estimating cause specific hazard functions for time-to-event predictions across two events. Note that hidden states in the ODE-RNN encoder (in green) as well as the patient’s latent health trajectory (in orange) have continuous dynamics specified by neural networks.}  
\end{figure*}

Following the recently proposed latent-variable time series model \citep{Chen2018, Rubanova2019}, we adopt a generative, variational autoencoder framework \citep{Kingma2013} to model time-varying features. As shown in Fig. \ref{fig:model_architecture_simple}, SurvLatent ODE encodes a patient-specific temporal trajectory of features with a varying time length, denoted by $\mathcal{X}_i$, into the latent embedding with a fixed dimension via the ODE-RNN encoder. Then, it decodes the patient specific embedding into the latent trajectory with pre-specified length $t_{m}$ (i.e. $Z_i^{t_m} = (z_{i,0}, z_{i,1}, ...,z_{i,t_m})$) by evaluating the integral with $g_{\phi}(\cdot)$ as the integrand, where $g_{\phi}(\cdot)$ is parameterized by the neural networks and describes the latent dynamics. Finally, we utilized cause-specific decoder modules, which consists of fully connected neural networks, to estimate hazard function for each event over time. 

\begin{figure*}[!ht]
  \centering
  \includegraphics[width=1\textwidth]{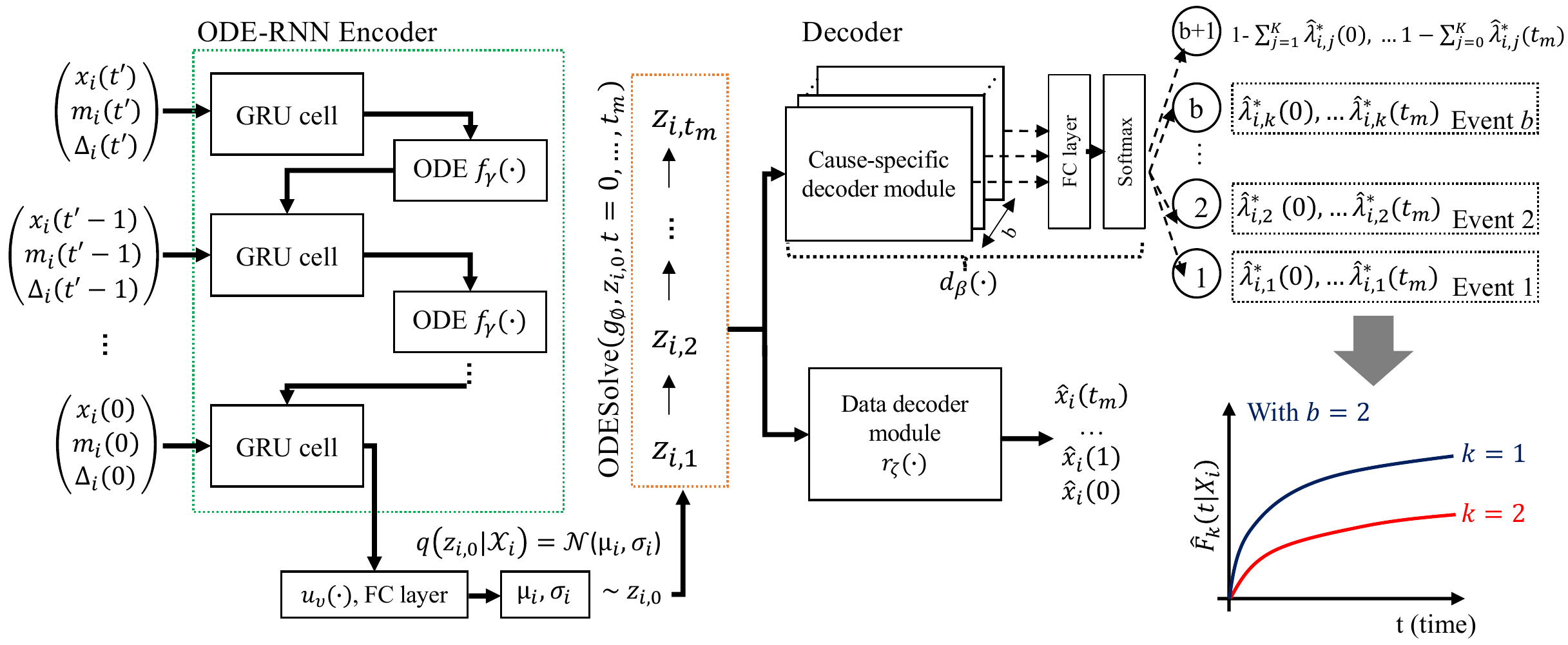}
  
\caption{\label{fig:model_architecture} \small \textbf{Detailed model architecture of SurvLatent ODE}. The ODE-RNN encoder takes a patient-specific time-varying features $\mathcal{X}_i$ along with missing pattern information defined over the union of discrete time points $(0, 1, ..., t'-1, t')$ and parameterizes approximate posterior over the initial latent variable $z_{i,0}$. A black-box differential equation solver ODESolve$(g_{\phi}, z_{i,0}, t = 0, ...,t_m)$ is then called over the sampled initial latent variable $z_{i,0}$ to obtain the latent trajectory, $Z_i^{t_m}  = (z_{i,0}, z_{i,1}, ..., z_{i,t_m})$. Finally, $d_\beta(\cdot)$, which consists of cause-specific decoder modules, subsequent fully connected layer, and softmax layer, maps the corresponding latent trajectory to estimate cause-specific hazard function for each event.}  
\end{figure*}

The detailed model architecture of SurvLatent ODE is shown in Fig. \ref{fig:model_architecture}. On the encoder side, we utilized an ODE-RNN \citep{Rubanova2019} to learn the latent dynamics of the input temporal data and parameterize approximate posterior over $z_0$ (i.e. $q(z_0|\mathcal{X})$). A function $f_{\gamma}(\cdot)$, parameterized by neural networks with $\gamma$ as trainable weights, specifies the dynamics of RNN hidden states, which allows the RNN to have continuously well-defined hidden states and incorporate inputs arriving at arbitrary times. This is the key element in the ODE-RNN encoder, which makes it unnecessary to explicitly impute missing measurements. When feeding the data into the model, we zero-fill missing measurements and concatenate the vector of indicators for missing measurements, $m(\cdot)$, as well as time elapsed since each feature's last observation, $\Delta(\cdot)$ \citep{Che2018Rnn}, into the input data $x(\cdot)$ over the union of all time points in each batch. 

For a RNN model, we utilize a Gated Recurrent Unit (GRU) cell \citep{Cho2014}, which resolves the vanishing or exploding gradient problems of conventional RNNs and enables learning long-term temporal dependencies across the data. Given the set of longitudinal data $\mathcal{X}$ with the latest measurement time of $t'$, the ODE-RNN encoder runs backwards in time from $t'$ to $0$ to get the approximate posterior $q(z_0|\mathcal{X})$ as follows:
\begin{equation}
    \begin{split}
        q(z_0|\mathcal{X}) &= \mathcal{N}(\mu_{z_0}, \sigma_{z_0}) \\
        \mu_{z_0}, \sigma_{z_0} &= u_{\upsilon}(\text{ODE-RNN$_{\theta, f_{\gamma}(\cdot)}$}(\mathcal{X})),
    \end{split}
\end{equation} where $\theta$ corresponds to a set of trainable weights in the GRU cell and $u_{\upsilon}$ is a neural network with weights $\upsilon$ and relates the final hidden state of the ODE-RNN encoder to the mean and variance over $z_{0}$. As a result of the ODE-RNN encoder, temporal latent dynamics of the hidden states are flexibly modeled by neural networks without resorting to any strong parametric assumptions on the dynamics \citep{Rubanova2019}.

On the decoder side, we utilize a multi-task learning framework \citep{Caruana1997, Alaa2017, Lee2020DynamicDeepHit} and allow for the latent trajectory $Z^t$ to be shared across multiple events. For patient $i$, the latent trajectory $Z_i^t$ is obtained by sampling the initial state $z_{i,0}$ from the approximate posterior $q(z_{i,0})$ and evaluating the integral, $z_{i,0} + \int_{0}^{t} g_{\phi}(z_{i,u}) du$ at pre-specified time points (i.e. $t = 0, ..., t_{m}$, where $t_{m}$ is the end time point of the prediction window and may be arbitrarily far into the future). This integral can be estimated by calling a black-box differential equation solver, ODESolve$(g_{\phi}, z_{i,0}, t = 0, ...,t_m)$ \citep{Chen2018}. Then, each cause-specific decoder module out of total $b$ modules learns a flexible event-specific function which maps the latent trajectory $Z^t$ shared across $b$ events to the cause-specific hazard function for each event $\lambda_k(t)$ (see Section \ref{ssec:survival_func_est}). This operationalizes the intuition that multiple clinical outcomes can have partially shared latent mechanisms. For example, the clinical outcomes such as VTE, cancer-specific death, and metastasis can have a shared latent mechanism driven by types of malignant tumors and blood protein tests. Finally, the data decoder module reconstructs a patient-specific data trajectory across the pre-specified time window (i.e. $\hat{x}_i(0), \hat{x}_i(1), ..., \hat{x}_i(t_{m})$), which could potentially be used for interpolation and extrapolation tasks. 

\subsubsection{Survival function estimation}
\label{ssec:survival_func_est}

Survival function estimation in our model comes down to learning the function, $d_\beta(\cdot)$, which maps the sampled patient-specific latent trajectory, $Z_i^{t_m}$ to the cause-specific hazard function for event $k$, $\lambda_{i,k}^*(t)$ evaluated across the pre-specified time window $[0, t_{m}]$. Then, we utilize the hazard function to estimate event free survival probability as well as cause-specific cumulative incidence function over the appropriate time window. $\beta$ corresponds to a set of trainable weights in cause-specific decoder modules and the subsequent fully-connected layer as shown in Fig. \ref{fig:model_architecture}. Recall that in our discrete time-to-event analysis framework, $\lambda_{i,k}^*(t) = P(T^{r} = t, K = k | T^{r} \geq t, \mathcal{X}_i, Z_i^t)$ where the hazard remains fixed in each time; bin widths are specified based on domain knowledge (e.g. a day or an hour). Note that our model does not specify any functional forms for how the data $\mathcal{X}$ relate to the $\lambda_{k}^*(t)$ and allow the neural networks to flexibly parameterize hazard functions. With the assumption that data ($\mathcal{X}$) is independent of remaining time-to-event ($T^{r}$) conditioning on the sampled latent state ($Z^t$), the event free survival survival probability $S(t|\mathcal{X}_i, Z_i^t)$ for patient $i$ is estimated using the corresponding latent trajectory $Z_i^t$ (i.e. patient $i$'s latent trajectory up until $t$) as follows:
\begin{equation}
    \label{eqn:ef_surv}
    \hat{S}(t|\mathcal{X}_i, Z_i^t) = \hat{S}(t|Z_i^t) = \hat{P}(T^{r} > t|Z_i^t) = \prod_{\tau_{i,l} < \tau \leq t} \bigg(1 - \sum_{k = 1}^b \hat{\lambda}_{i, k}^*(\tau)\bigg),
\end{equation} where $\tau_{i,l}$ is the latest measurement time for the patient. Setting a lower-bound for $\tau$ with $\tau_{i,l}$ is necessary since $S(t|Z_i^t)$ should always be 1 for $t \in [0, \tau_{i,l}]$. The proof for Equation \ref{eqn:ef_surv} is shown in Appendix \ref{appendix:appendix_a_discret}. Notice that at any given time point $\tau$, $\sum_{k = 0}^b\lambda_{i, k}^*(\tau)$ must sum up to 1, where $\lambda_{i, 0}^*(\tau) = 1 - \sum_{k = 1}^b\lambda_{i, k}^*(\tau)$ and denotes probability of not experiencing any events at time $\tau$ given patient $i$ survived up to the prior time point (i.e. $\tau - 1$). We enforce this constraint using the softmax layer as shown in Fig. \ref{fig:model_architecture}. Finally, the estimated cause-specific cumulative incidence function $\hat{F}_k(t|Z_i^t)$ is
\begin{equation}
     \hat{F}_k(t|Z_i^t) = \hat{P}(T^r \leq t, K = k|Z_i^t) = \sum_{\tau_{i,l} < \tau \leq t} \hat{P}(T^r = \tau, K = k|Z_i^t) = \sum_{\tau_{i,l} < \tau \leq t} \hat{\lambda}_{i,k}^*(\tau)\hat{S}(\tau - 1|Z_i^t).
\end{equation}

\subsection{Inference}
\label{sec:inference}

We devise the loss function which handles time-varying features as well as loss to follow-up (i.e. right-censoring). The total loss consists of two components : the Kullback-Leibler (KL) divergence loss (i.e. $D_{KL}[q(z_0|\mathcal{X})~ || ~p(z_0 | \mathcal{X})]$), where $q(z_0)$ is the approximate posterior over the initial latent state and $\mathcal{X}$ is the set of time-varying features, and the log of the total survival likelihood. In practice, the KL loss cannot be directly evaluated since the posterior distribution $p(z_0|\mathcal{X})$ is intractable because computing evidence $p(\mathcal{X})$ involves marginalizing out high-dimensional latent variable $z_0$. Therefore, we instead maximize the evidence lower bound (ELBO) :
\begin{equation}
    \begin{split}
        \text{ELBO}(\mathcal{X};\Phi, \zeta) &= \E_{q(z_0|\mathcal{X};\Phi)} [\text{log}(p(\mathcal{X}|z_0;\phi, \zeta)] - \text{KL}[q(z_0|\mathcal{X};\Phi)~||~p(z_0)],\\
    \end{split}
\end{equation} where $p(z_0)$ is assumed to be a standard normal, $\Phi$ corresponds to a set of trainable weights $\{\theta, \gamma, \upsilon, \phi\}$ in the ODE-RNN encoder and $g_{\phi}(\cdot)$ which specifies dynamics of the latent trajectory on the decoder side, and $\zeta$ is a set of trainable weights in the data decoder module, $r_{\zeta}(\cdot)$. Maximizing the ELBO in this way is equivalent to minimizing the KL divergence \citep{Bishop2006}.

The total survival likelihood $L_{\text{surv}}$, which enables handling of right-censored patients, is estimated as follows:
\begin{equation}
    \begin{split}
        L_{\text{surv}}(\mathcal{D}; \Phi, \beta) = \prod_{i \in \mathcal{D}}&\hat{P}(T^{r} = t_i^{r}, K = k_i;\Phi, \beta)^{\delta_i} \times \hat{P}(T^{r} > t_i^{r};\Phi, \beta)^{1 - \delta_i}\\
        = \prod_{i \in \mathcal{D}}& \Big[\hat{\lambda}_{i,k}^*(t_i^{r};\Phi, \beta) \hat{S}(t_i^{r} - 1|X_i;\Phi, \beta)\Big]^{\delta_i}\times \hat{S}(t_i^{r}|X_i;\Phi, \beta)^{1 - \delta_i},
    \end{split}
\end{equation} where $t_i^{r}$ is the observed remaining follow-up time from the latest measurement, $k_i$ is the observed event type of patient $i$, $\delta_i$ is an event indicator, and $\beta$ is a set of trainable weights in cause-specific decoder modules and the subsequent fully connected neural network (i.e. $d_\beta(\cdot)$). Note that the likelihood contribution from a right-censored patient (i.e. those with $\delta_i = 0$) is that they are alive at $t_i^{r}$, which corresponds to the event free survival probability $S(t_i^{r}|Z_i)$ (see Equation \ref{eqn:ef_surv}).

Therefore, the total loss we want to minimize is 
\begin{equation}
    \begin{split}
        L_{\text{total}}(\mathcal{D};\Phi, \zeta, \beta) = - \text{ELBO}(\mathcal{X};\Phi, \zeta) - \text{log}(L_{\text{surv}}(\mathcal{D}; \Phi, \beta))
    \end{split}.
\end{equation} We perform an end-to-end training and utilize a standard back-propagation method to update weights in each module. For ODE solvers, we use the fifth order Dormand–Prince method \citep{Dormand1980} from \verb|torchdiffeq| Python package \citep{Chen2018}. Alternatively, the adjoint method, proposed in \cite{Chen2018}, can be used to compute more memory-efficient gradients for ODE solvers, but at a cost of longer computation time. 

\section{Experiments}
\subsection{Setup}
To demonstrate clinical utility of our model, we evaluated SurvLatent ODE on two real-world observational datasets: MIMIC-III for predicting time to hospital mortality (i.e. single-outcome) and the data from the Dana-Farber Cancer Institute (DFCI) for predicting time to VTE with all-cause mortality as a competing event. The datasets include routinely collected longitudinal measurements of patients' biomarkers where some measurements are often informatively missing. For both prediction tasks, patients were aligned at the first measurement time upon their admission. Finally, we randomly split each data into a training set (55\%), validation set (15\%), and test set (30\%), tuned hyperparameters using train and validation sets (see Section \ref{ssec:hyperparam}), and obtained model performances on the held-out test set across 25th, 50th, and 75th percentiles of corresponding event times. 

\subsection{Datasets}
\paragraph{MIMIC-III}

MIMIC-III is a large, publicly available longitudinal dataset of patients who were admitted to critical care units at the Beth Israel Deaconess Medical Center in Boston, Massachusetts \citep{Johnson2016}. In critical care scenarios, it is important to accurately identify high-risk patients as well as patient-specific time to mortality based on their time-varying biomarkers. This allows clinicians to determine the right level of care for patients in a timely manner based on their estimated risks and improve health outcomes while reducing unnecessary resource utilization. 

For predicting time to hospital mortality, we utilized the cohort which consists of patients who were at risk after the first 36 hours of the admission (n = 21,728) and estimated their remaining time to mortality based on a total of 42 features. The features include static features such as age and sex as well as time-varying features such as vital signs and laboratory tests across the first 36 hours at a time resolution of 1 hour. For the longitudinal features, we included the top 40 most frequent vital signs and laboratory test results (see Appendix \ref{appendix:model_detail} for the full set of features). As an outcome, we used the time to hospital mortality measured from the 36th hour while right-censoring those who did not experience the mortality at their hospital discharge time. Among the total of 21,728 patients in our cohort, 2,281 patients (10.5 \%) died at the hospital. 

\paragraph{Dana-Farber Cancer Institute (DFCI) dataset}
Venous Thromboembolism (VTE) is a frequent, yet fatal complication in patients with active cancer, especially while they are receiving chemotherapy. VTE is associated with worse clinical outcomes such as elevated risk of mortality and reduced quality of life, and may lead to a significant disturbance of cancer treatment regimens \citep{Blom2006, Khorana2007, Lloyd2018}. For ambulatory cancer patients, preventive anti-thrombotic measures such as thromboprophylaxis, a mechanical method to promote venous outflow, and anticoagulants drugs are effective at reducing the incidence of VTE \citep{Rutjes2020, Key2020, Xiong2021}, however with an increased risk of bleeding \citep{Key2020}. Therefore, accurate stratification of the VTE risk among patients with cancer may allow clinicians to improve clinical outcome while minimizing side effects due to overtreatment. 

The dataset includes patients with active cancer who were admitted to the DFCI from 2/12/2013 to 12/14/2021 and received targeted panel sequencing of their tumor biopsy. We set the sequence date for each patient as the prediction time to avoid immortal time bias. For the experiment cohort, we included those who received the sequencing within a year of their admission and were at the risk of VTE without any recorded prior VTE event (n = 8,734). The total number of features is 64, which includes static features such as age, sex, presence of metastasis, and diagnosed cancer types as well as time-varying features such as laboratory test results and body mass index (BMI) across the first year from admission at a time resolution of one day (see Appendix \ref{appendix:model_detail} for the full set of features). We utilized ICD-10 diagnosis codes, I82.4 and I82.6, to determine the time of the venous thromboembolism (VTE) onset and I26 to determine the time of pulmonary embolism (PE) onset. And, we combined VTE and PE onset as the VTE event. We used the National Death Index (NDI) up until 12/31/2020 and the in-house death registry after the date to determine time of mortality. Finally, we right-censored patients at their last hospital visits, who did not experience VTE and death events. The demographics summary of the chosen cohort as well as causes of death (Table \ref{tab:dvt_cohort_summary} and \ref{tab:cause_of_death_in_data}) are shown in Appendix \ref{appendix:dvt_detail}.

The main goal of this experiment was to develop a risk stratification framework for cancer-related VTE by accurately estimating the time to VTE event in the presence of a competing risk for death as well as missing measurements and loss to follow-ups. We thus estimated a patient specific remaining time to VTE from the sequence date with all-cause mortality as a competing event. Finally, to demonstrate clinical utility of the proposed model, we benchmarked against the categorical Khorana score \citep{Khorana2008}, which is the current clinical standard for cancer-related VTE stratification and utilizes four clinical measurements together with cancer type (see Section \ref{ssec:baselines} for details).

\subsection{Baselines}
\label{ssec:baselines}

\subsubsection{Time to hospital mortality prediction (MIMIC-III)}
\textbf{Surv VAE-RNN}, modified from \citep{Che2018Rnn}, adopts a variational autoencoder framework where both encoder and decoder are recurrent neural networks. This approach is closest to the proposed model except that in our model we utilized deep neural networks to learn the latent state dynamics.

\textbf{Recurrent Deep Survival Machine (RDSM}, \cite{Chirag2021DeepParam}) is a fully parametric time-to-event model where it learns representation of the input longitudinal data using a RNN-based model and derives the conditional survival distribution (i.e. $P(T^r = t|X)$) as a fixed mixture of parametric distributions (e.g. Weibull or Log-Normal).

\textbf{Dynamic-Deephit} \citep{Lee2020DynamicDeepHit} is a discrete time-to-event model which, like RDSM, learns input representation using a RNN-based model and estimates the conditional joint distribution of the survival time and event type (i.e. $P(T = t, K = k|X)$).

\textbf{Cox Proportional Hazard model (Cox PH)}, proposed by \cite{Cox1972}, is a popular semi-parametric, time-to-event model which relates log of the relative hazard to a linear function of baseline features. For the experiment, we forward-filled the missing measurements to obtain the baseline features.

Refer to Table \ref{tab:model_comparison_attrs} for the summary of the baseline models as well as the proposed model in terms of key strategies in handling longitudinal data.
\begin{table}[!ht]
\centering
\resizebox{0.9\textwidth}{!}{%
\begin{tabular}{cccccccc}
 &
  \multirow{2}{*}{\begin{tabular}[c]{@{}c@{}}Handles\\ time-varying\\ features\end{tabular}} &
  \multicolumn{1}{c|}{\multirow{2}{*}{\begin{tabular}[c]{@{}c@{}}Handles\\ Competing\\ risks\end{tabular}}} &
  \multicolumn{2}{c|}{\textit{Handling missing measurements}} &
  \multicolumn{2}{c|}{\textit{Learning latent state dynamics}} &
  \multirow{2}{*}{\begin{tabular}[c]{@{}c@{}}Generative\\ model (VAE)\end{tabular}} \\
 &
   &
  \multicolumn{1}{c|}{} &
  \begin{tabular}[c]{@{}c@{}}Include missing\\ pattern in input\end{tabular} &
  \multicolumn{1}{c|}{\begin{tabular}[c]{@{}c@{}}Mean imputation/\\ forward-filling\end{tabular}} &
  \multicolumn{1}{c}{\begin{tabular}[c]{@{}c@{}}Exponential\\ decay\end{tabular}} &
  \multicolumn{1}{c|}{\begin{tabular}[c]{@{}c@{}}Neural\\ Networks\end{tabular}} &
   \\ \Xhline{3\arrayrulewidth}
\begin{tabular}[c]{@{}c@{}}\textbf{SurvLatent ODE}\\ \textbf{(Proposed model)}\end{tabular} & V & V           & V &   &   & V & V \\ \hline
\begin{tabular}[c]{@{}c@{}}Surv RNN-VAE\\ (Modified from \cite{Che2018Rnn})\end{tabular}         & V &             & V &   & V &   & V \\ \hline
\begin{tabular}[c]{@{}c@{}}RDSM\\ \citep{Chirag2021DeepParam}\end{tabular}                 & V &             &   & V &   &   &   \\ \hline
\begin{tabular}[c]{@{}c@{}}Dynamic-Deephit\\ \citep{Lee2020DynamicDeepHit}\end{tabular}      & V & V           & V &   &   &   &   \\ \hline
\begin{tabular}[c]{@{}c@{}}Cox PH\footnotemark\\ \citep{Cox1972}\end{tabular}               &   &  &   & V &   &   &   \\ \Xhline{3\arrayrulewidth}
\end{tabular}%
}
\caption{\small Model comparison in terms of key strategies in handling longitudinal data. SurvLatent ODE is a generative time-to-event framework which flexibly models latent state dynamics using neural networks, which sets it apart from other recent work.}
\label{tab:model_comparison_attrs}
\end{table}

\footnotetext[1]{Variations of Cox PH framework have been proposed for handling time-varying features \citep{Fisher1999} and competing risks \citep{Fine1999, Putter2007}.}

\subsubsection{Time to VTE prediction with death as a competing event (DFCI data)}
\textbf{Khorana score} \citep{Khorana2008} is a simple linear scoring rule widely used in hospitals to predict future risk of VTE for patients with cancer. The score ranges from 0 to 6 and is computed based on the patient's cancer type, body mass index (BMI), and lab test results including pre-chemotherapy platelet count, hemoglobin level, and pre-chemotherapy leukocyte count. 1,031 patients among the test cohort (n = 2,630) have the relevant features to compute Khorana scores at the prediction time. To obtain Khorana scores for the rest of patients in the test cohort, we imputed the features for those without BMI measurements and/or the relevant lab test results at the prediction time using population means.

\textbf{Cause-specific Cox model (CS Cox)} is a Cox PH based model for handling competing risks and captures effects of the static baseline features on the cause-specific hazard function for each event \citep{Putter2007}. Similar to the Cox PH setting above, we forward-filled the missing values at the prediction time.  

\textbf{Fine-Gray Cox model (FG Cox)}, proposed by \cite{Fine1999}, is another Cox PH based model, widely used for competing risks. The model directly captures effects of the baseline features on the cumulative incidence function to make survival predictions. We adopted the missing measurement imputation strategy exactly identical to CS Cox.

\subsection{Evaluation Metrics}
\label{ssec:eval_metrics}

\textbf{Time-dependent cumulative/dynamic AUC} corresponds to the probability that, given a pair of patients where one experienced the event of interest before $t$ and the other is event-free at $t$, the model correctly ranks their risks of the event \citep{Kamarudin_2017}.

\textbf{Time-dependent Brier Score} is a generalization of the conventional Brier score \citep{Brier1950} which can account for the right-censored data and measures the mean squared error at a given time point $t$. This metric evaluates the model's calibration as well as discrimination performance \citep{Xiu2020}. 

See Appendix \ref{appendix:eval_metrics} for details on the evaluation metrics.

\subsection{Hyperparameters}
\label{ssec:hyperparam}
We used the validation set for optimizing hyperparameters and utilizing early-stopping to avoid over-fitting. We utilized a random search \citep{Bergstra2012}, with the set of hyperparameters including the number of dimensions in the encoder and decoder networks, the number of nodes in RNN hidden layers, the number of hidden layers and nodes in fully-connected neural networks for ODEs. For a fair comparison, we similarly tuned the deep learning baseline models including Surv RNN-VAE, RDSM, and Dynamic-Deephit. See Appendix \ref{appendix:model_detail} for the full list of hyperparameters and their possible values across all models.

\section{Results on Real Data} 
\subsection{Results on MIMIC-III} 
\begin{table*}[!ht]
\centering
\resizebox{\textwidth}{!}{%
\begin{tabular}{llll|lll}
 &
 \multicolumn{3}{c}{Time-dependent AUC$(t)$} &
 \multicolumn{3}{c}{Brier Score, $\text{BS}(t)$} \\
 &
  \begin{tabular}[c]{@{}l@{}}25th percentile\\(Hour 35)\end{tabular} &
  \begin{tabular}[c]{@{}l@{}}50th percentile\\(Hour 81)\end{tabular} &
  \begin{tabular}[c]{@{}l@{}}75th percentile\\(Hour 150)\end{tabular} &
  \begin{tabular}[c]{@{}l@{}}25th percentile\\(Hour 35)\end{tabular} &
  \begin{tabular}[c]{@{}l@{}}50th percentile\\(Hour 81)\end{tabular} &
  \begin{tabular}[c]{@{}l@{}}75th percentile\\(Hour 150) \end{tabular} \\ \Xhline{3\arrayrulewidth}

\textbf{\begin{tabular}[c]{@{}l@{}}SurvLatent ODE \\ (Proposed model)\end{tabular}} &
  \textbf{0.920 (0.009)} &
  \textbf{0.883 (0.009)} &
  \textbf{0.831 (0.010)} &
  \textbf{0.0220 (0.0013)} &
  \textbf{0.0442 (0.0019)} &
  0.0789 (0.0029) \\ \hline
Surv RNN-VAE &
  0.535 (0.022)$^{**}$ &
  0.535 (0.016)$^{**}$ &
  0.521 (0.014)$^{**}$ &
  0.0281 (0.0017)$^{**}$ &
  0.0571 (0.0023)$^{**}$ &
  0.0950 (0.003)$^{**}$ \\ \hline
RDSM &
  0.836 (0.017)$^{**}$ &
  0.817 (0.013)$^{**}$ &
  0.784 (0.011)$^{**}$ &
  0.0241 (0.0018)$^{*}$ &
  0.0449 (0.0023) &
  \textbf{0.0618 (0.0025)} \\ \hline
Dynamic-Deephit &
  0.891 (0.009)$^{**}$ &
  0.860 (0.009)$^{*}$ &
  0.808 (0.010)$^{*}$ &
  0.0247 (0.0018)$^{*}$ &
  0.0492 (0.0024)$^{**}$ &
  0.0816 (0.0032) \\ \hline
Cox PH &
  0.826 (0.017)$^{**}$ &
  0.806 (0.013)$^{**}$ &
  0.762 (0.012)$^{**}$ &
  0.0234 (0.0017) &
  0.0465 (0.0023) &
  0.0789 (0.0032) \\ \Xhline{3\arrayrulewidth}
\end{tabular}%
}
\caption{\small Performance summary of the models across the 25th, 50th, and 75th percentiles of event times on the held-out test set of MIMIC-III data (n = 6,519) for predicting \textbf{time to hospital mortality}. Time-dependent AUC (higher the better) as well as Brier scores (lower the better) are used as evaluation metrics (see Section \ref{ssec:eval_metrics}). Standard errors, obtained from the non-parametric bootstrap on the test cohort, are shown in parenthesis. We obtained statistical significance ($* : p < 0.05, ** : p < 0.001$) by estimating mean of differences in performance between the proposed model (SurvLatent ODE) and each baseline model across bootstrap iterations and performing a one-sided test.}
\label{tab:mimic_perf}
\end{table*}

\begin{figure}[!htb]
  \centering
  \includegraphics[width=0.60\textwidth]{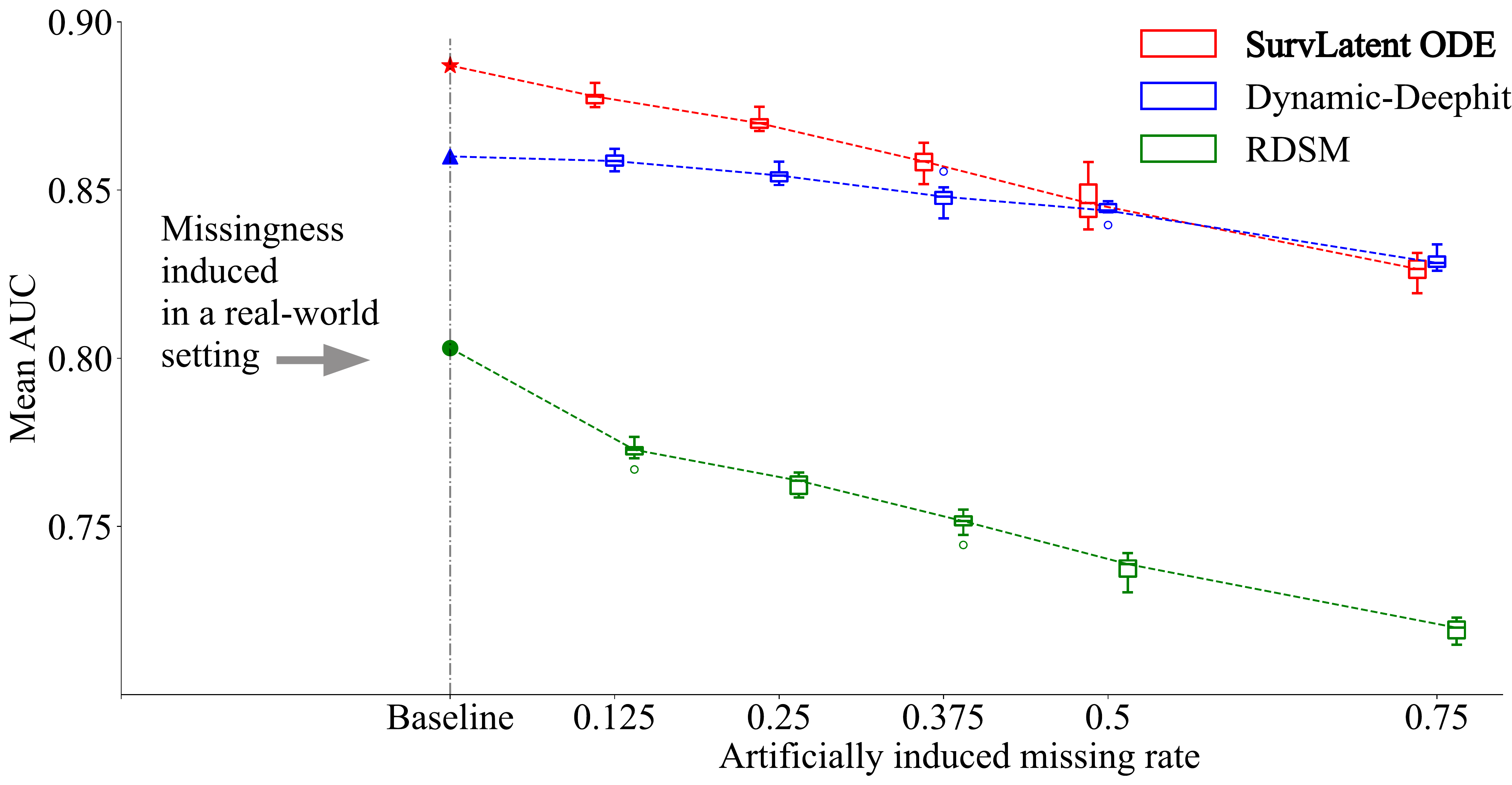}
  \centering

\caption{\label{fig:missing_eval} \small Performance of SurvLatent ODE and deep learning based models across increasing missing rate over the test set from the MIMIC-III dataset. Measurements under randomly chosen time points per each patient were dropped. At each missing rate, a box plot represents the performance of each model evaluated across 10 different subsampled versions of the test set. To obtain a single summary measure for each evaluation, we averaged time-dependent AUC over 25th to 75th percentiles of the event times.}
\end{figure}

As shown in Table \ref{tab:mimic_perf}, for the MIMIC dataset experiment, SurvLatent ODE significantly outperformed the baseline models including recently published deep learning based approaches across all event time percentiles in the discriminative metric, while showing competitive performance in calibration captured by the Brier Score. The Cox PH model had the lowest discriminative performance, as it enforced a linear relationship between the set of features and did not incorporate the evolution of time-varying features. As an ablation study, we noticed that Surv RNN-VAE, which closely resembles the proposed model except how latent state dynamics are modeled (i.e. parameterized exponential decay vs. neural networks), showed fairly poor performance across all metrics. This demonstrates that a Neural-ODE based approach allows the variational autoencoder framework to be successfully implemented for time-to-event modeling of longitudinal data, which would have been infeasible otherwise.

SurvLatent ODE demonstrated competitive and robust performance over randomly subsampled test data with increasing missing rate (Fig. \ref{fig:missing_eval}). To investigate each model's generalizability to more sparse patients' data, we evaluated the best model from the earlier experiment (Table \ref{tab:mimic_perf}) on 10 different randomly subsampled versions of the test set at each missing rate. At the baseline, each patient data in the test set already contains substantial missing measurements ($\sim$ 70\% on average) across the first 36 hours of admission and 40 time-varying features, driven by real-world mechanisms. RDSM, which relies on mean imputation for handling missing measurements, showed relatively poor performance across the induced missing rates. While Dynamic-Deephit showed consistent performance across the missing rates, SurvLatent ODE, with the substantial performance gain at the baseline real-world condition, continued to outperform all the other models through the artificially induced missing rate of 0.5. 

\subsection{Results on VTE} 
\begin{table*}[!ht]
\centering
\resizebox{\textwidth}{!}{%
\begin{tabular}{llll|lll}
  &
  \multicolumn{3}{c}{Time-dependent AUC$_k(t)$} &
  \multicolumn{3}{c}{Brier Score, $\text{BS}_k(t)$} \\
  &
  \begin{tabular}[c]{@{}l@{}}25th percentile\\(Day 47)\end{tabular} &
  \begin{tabular}[c]{@{}l@{}}50th percentile\\(Day 113)\end{tabular} &
  \begin{tabular}[c]{@{}l@{}}75th percentile\\(Day 266)\end{tabular} &
  \begin{tabular}[c]{@{}l@{}}25th percentile\\(Day 47)\end{tabular} &
  \begin{tabular}[c]{@{}l@{}}50th percentile\\(Day 113)\end{tabular} &
  \begin{tabular}[c]{@{}l@{}}75th percentile\\(Day 266)\end{tabular} \\ \Xhline{3\arrayrulewidth}
\begin{tabular}[c]{@{}l@{}}\textbf{SurvLatent ODE}\\ \textbf{(Proposed model)}\end{tabular} &
  \textbf{0.782 (0.031)} &
  \textbf{0.781 (0.021)} &
  \textbf{0.758 (0.020)} &
  \textbf{0.0222 (0.0026)} &
  \textbf{0.0426 (0.0034)} &
  \textbf{0.0631 (0.0040)}\\ \hline
\begin{tabular}[c]{@{}l@{}} Dynamic-Deephit\end{tabular} &
  0.729 (0.032)$^{*}$ &
  0.770 (0.023) &
  0.722 (0.022)$^{*}$ &
  0.0223 (0.0027) &
  0.0428 (0.0036) &
  0.0641 (0.0042)$^{*}$\\ \hline
CS Cox &
   0.702 (0.034)$^{*}$ &
   0.728 (0.023)$^{*}$ &
   0.706 (0.020)$^{*}$ &
   0.0228 (0.0028) &
   0.0444 (0.0038)$^{*}$ &
   0.0666 (0.0045)$^{**}$ \\ \hline
FG Cox &
   0.672 (0.033)$^{*}$ &
   0.695 (0.024)$^{*}$ &
   0.686 (0.019)$^{**}$ &
   0.0227 (0.0028) &
   0.0445 (0.0038)$^{*}$ &
   0.0667 (0.0045)$^{**}$ \\\Xhline{3\arrayrulewidth}
Khorana scores\footnotemark &
  0.625 (0.051) &
  0.592 (0.036) &
  0.581 (0.029) &
  N/A &
  N/A &
  N/A \\ 
\begin{tabular}[c]{@{}l@{}}Khorana scores \\ (imputed)\end{tabular} &
  0.627 (0.036)$^{**}$&
  0.628 (0.026)$^{**}$&
  0.628 (0.022)$^{**}$&
  N/A &
  N/A &
  N/A  \\ \Xhline{3\arrayrulewidth}
\end{tabular}%
}
\caption{\small Performance summary of the models across the 25th, 50th, and 75th percentiles of event times on the held-out test set of the in-house DFCI data (n = 2,630) for predicting \textbf{time to VTE event} with an all-cause mortality as a competing event. Time-dependent AUC (higher the better) as well as Brier scores (lower the better) are used as evaluation metrics. Standard errors, shown in parenthesis, as well as statistical significance ($* : p < 0.05, ** : p < 0.001$) were obtained exactly in the same manner as in the MIMIC-III experiment (see Table \ref{tab:mimic_perf} caption). Note that Khorana score adopts the integer scoring system and its survival calibration performance cannot be evaluated.}
\label{tab:dvt_perf}
\end{table*}

\begin{table*}[!h]
\centering
\resizebox{\textwidth}{!}{%
\begin{tabular}{llll|lll}
  &
  \multicolumn{3}{c}{Time-dependent AUC$_k(t)$} &
  \multicolumn{3}{c}{Brier Score, $\text{BS}_k(t)$} \\
  &
  \begin{tabular}[c]{@{}l@{}}25th percentile\\(Day 141)\end{tabular} &
  \begin{tabular}[c]{@{}l@{}}50th percentile\\(Day 269)\end{tabular} &
  \begin{tabular}[c]{@{}l@{}}75th percentile\\(Day 446)\end{tabular} &
  \begin{tabular}[c]{@{}l@{}}25th percentile\\(Day 141)\end{tabular} &
  \begin{tabular}[c]{@{}l@{}}50th percentile\\(Day 269)\end{tabular} &
  \begin{tabular}[c]{@{}l@{}}75th percentile\\(Day 446)\end{tabular} \\ \Xhline{3\arrayrulewidth}
\begin{tabular}[c]{@{}l@{}}\textbf{SurvLatent ODE}\\ \textbf{(Proposed model)}\end{tabular} &
  \textbf{0.772 (0.018)} &
  \textbf{0.762 (0.014)} &
  \textbf{0.761 (0.012)} &
  0.0785 (0.0028) &
  0.1270 (0.0031) &
  0.1626 (0.0032) \\ \hline
\begin{tabular}[c]{@{}l@{}} Dynamic-Deephit\end{tabular} &
  0.762 (0.018) &
  0.742 (0.014)$^{*}$ &
  0.741 (0.012)$^{**}$ &
  \textbf{0.0662 (0.0041)} &
  0.1198 (0.0047) &
  0.1600 (0.0044) \\ \hline
CS Cox &
   0.711 (0.020)$^{**}$ &
   0.710 (0.014)$^{**}$ &
   0.710 (0.013)$^{**}$ &
   0.0675 (0.0046) &
   0.1204 (0.0055) &
   0.1602 (0.0054) \\ \hline
FG Cox &
   0.706 (0.020)$^{**}$ &
   0.706 (0.014)$^{**}$ &
   0.706 (0.013)$^{**}$ &
   0.0667 (0.0045) &
   \textbf{0.1179 (0.0050)} &
   \textbf{0.1580 (0.0048)} \\\Xhline{3\arrayrulewidth}
\end{tabular}%
}
\caption{\small Performance summary of the models across the 25th, 50th, and 75th percentiles of event times for predicting \textbf{time to all-cause mortality} as the competing event. Performances for VTE (Table \ref{tab:dvt_perf}) and all-cause mortality (Table \ref{tab:death_perf}) were obtained from a single model evaluated on the same held-out test set of the in-house data (n = 2,630). Time-dependent AUC (higher the better) as well as Brier scores (lower the better) are used as evaluation metrics.}
\label{tab:death_perf}
\end{table*}

\footnotetext[2]{As mentioned in Section \ref{ssec:baselines}, this cohort includes 1,031 patients with recorded relevant lab test results and BMI to compute Khorana scores at the prediction time (i.e. panel sequencing data).}

As shown in Table \ref{tab:dvt_perf} and \ref{tab:death_perf}, SurvLatent ODE significantly outperformed all the baseline models including Dynamic-Deephit in the discriminative as well as the calibration metrics, with a statistically significant improvement over the next-best method in two out of three time horizons for the discriminative performance. For predicting time to all-cause mortality (i.e. the competing event), SurvLatent ODE again outperformed all the baseline models in the discriminative metric, with statistically significant improvement for two out of three time horizons, while achieving competitive but weaker performance in the calibration metric. Specifically, Dynamic-Deephit showed superior calibration at the short survival quantile (25\%) while the Fine-Gray Cox model (FG Cox) showed superior calibration for the longer survival quantiles (50\% and 75\%).  

SurvLatent ODE significantly outperformed the current clinical standard, Khorana scores \citep{Khorana2008} for predicting time to VTE events across all percentiles (Table \ref{tab:dvt_perf}). Khorana scores range from 0 to 6 and are meant to stratify risk for VTE among patients with cancer based on 5 different features including diagnosed cancer type and lab test results (see Section \ref{ssec:baselines} for details). The performance of Khorana scores was comparatively poor, even after the imputation. Notably, SurvLatent ODE continued to outperform the Khorana score when focusing on the highest risk cancer types for VTE, pancreatic and stomach cancer (Fig. \ref{fig:khorana_subcancer} in Appendix \ref{appendix:dvt_detail}). Furthermore, the proposed model demonstrated a strong performance across various subsets of patients defined by diagnosed cancer types and cancer stage (see Table \ref{tab:cancer_type_spec_perf} in Appendix \ref{appendix:dvt_detail}).

\begin{figure}[!htb]
  \centering
  \includegraphics[width=0.45\textwidth]{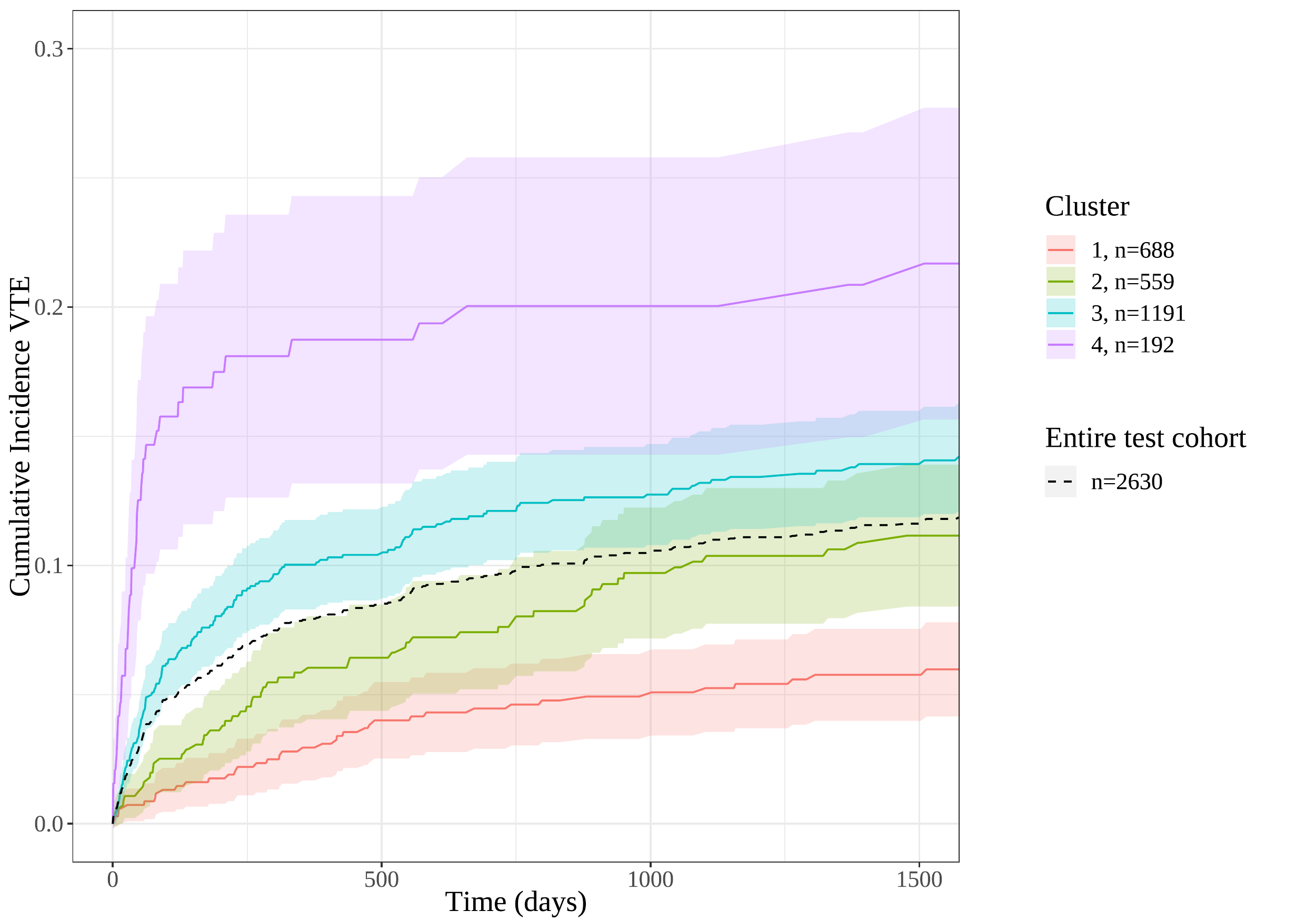}
  \centering
  \includegraphics[width=0.45\textwidth]{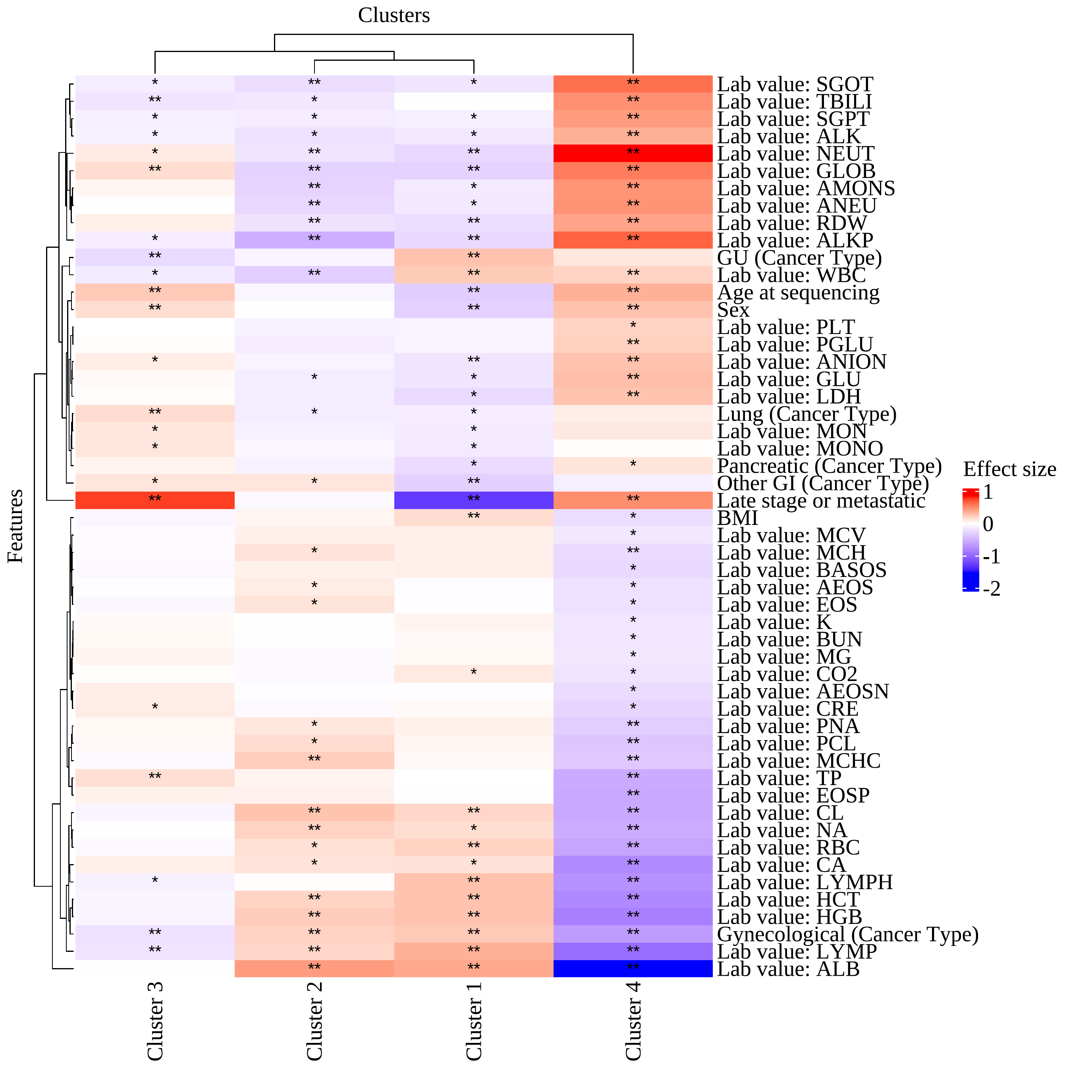}

\caption{\label{fig:latent_clustering_interpretation} \small \textbf{Left} : Cumulative Incidence trajectories for the VTE event across the latent clusters. The cumulative incidence trajectories were estimated by the Aalen-Johansen Estimator \citep{Johansen1978AnET}. \textbf{Right} : Heatmap visualization of feature-wise coefficient for each latent cluster. See Appendix \ref{appendix:dvt_detail} for detailed description of the method. SurvLatent ODE provided clinically meaningful and interpretable latent states. Features significantly associated with Cluster 4 include well-established risk factors for VTE. Note that $* : p < 0.05, ** : p < 0.001$ for statistical significance of feature coefficient.}
\end{figure}

As shown in Fig. \ref{fig:latent_clustering_interpretation}, SurvLatent ODE provided clinically meaningful and interpretable latent states with the obtained latent clusters (k = 4) exhibiting significantly different VTE risk trajectories. The latent state for each patient was obtained from the output of the cause-specific decoder for the VTE event (see Appendix \ref{appendix:dvt_detail} for more detail). We obtained a cluster specific coefficient for each baseline feature by running a logistic regression with cluster membership as the binary outcome. Khorana scores were also included as a categorical feature to highlight the contribution of features beyond the current clinical guidelines. Well-established VTE risk factors were associated with the highest risk cluster (i.e. Cluster 4), including cancer type (e.g. pancreatic cancer) and known lab test results (platelet count : PLT, hemoglobin level : HGB, and white blood cell count : WBC) with consistent directions \citep{Khorana2008}. While some of these factors are included in the Khorana score, the fact that they were still significantly associated after conditioning on the Khorana score indicates that SurvLatent ODE identified additional latent trajectory variations among these factors. In addition to the features for the Khorana score, we observed significant associations with neutrophils (NEUT), albumins (ALB), and SGOT which were supported by the literature \citep{Kapoor2018, Folsom2010, Folsom2014}.

\section{Discussion} 

We propose a full ODE-based variational autoencoder time-to-event framework for modeling longitudinal data, which encodes a patient-specific trajectory of time-varying features and decodes the resulting embedding into the latent trajectory for time-to-event prediction, where dynamics of the trajectories are learned by neural networks. Combined with the flexible estimation of cause-specific hazard functions for multiple events without explicit parametric assumptions on event time distributions, our proposed model outperforms the baselines including recently published deep-learning based models in two real-world datasets. Our proposed method offers a promising deep-learning based time-to-event framework for risk predictions in healthcare scenarios where data irregularities such as missing measurements and loss to follow-up are very common. 

Our model significantly outperforms the current clinical standard, Khorana scores, for predicting the VTE event among patients with cancer. In addition, by learning complex non-linear relationships between time-varying features, our model makes accurate time-to-event predictions for a wider range of patients while effectively capturing heterogeneity of the VTE risk, including identifying low-risk patients in conventionally high-risk cancer types. As future clinically-oriented work, we consider incorporating major fixed-effect variables, such as patient's tumor genetics, as well as more granular outcomes such as stroke, myocardial infarction, and cancer disease progression.

On the technical side, future work may involve estimating well-calibrated uncertainty of time-to-event predictions, reflective of training data availability for a given task as well as sample-wise prediction difficulty. In addition, as demonstrated in \cite{Garnelo18a, Norcliffe2021}, one can explicitly capture uncertainty over underlying dynamics and render time-to-event estimates aware of the uncertainty in the dynamics across latent states. From the time-to-event modeling perspective, it is of interest to additionally adopt the sub-distribution hazard framework which directly models effects of covariates on cause-specific cumulative incidence functions and is often used for an individual risk prediction \citep{Austin2016, Berger2020}.

\acks{This work was supported by NIH R01 CA227237, Louis B. Mayer Foundation, DDCF Clinical Scientist Development Award, Phi Beta Psi Sorority, and The Emerson Collective.}

\bibliography{reference}
\pagebreak
\appendix
\section{Discrete time-to-event analysis}
\label{appendix:appendix_a_discret}

In Section \ref{ssec:survival_func_est}, given that $T^r \indep \mathcal{X} | Z$, we claim the following 

\begin{align*}
    S(t|\mathcal{X}_i, Z_i^t) = P(T^{r} > t|\mathcal{X}_i, Z_i^t) = P(T^{r} > t|Z_i^t) = \prod_{\tau_{i,l} < \tau \leq t} \bigg(1 - \sum_{k = 1}^b \lambda_{i, k}^*(\tau)\bigg).
\end{align*}

\begin{proof}
For a discrete event time $T^r$, we notice that 
\begin{equation}\label{eqn:ef_surv_proof_0}
    S(t|Z_i^t) = \prod_{\tau_{i,l} < \tau \leq t} \frac{S(\tau | Z_i^t)}{S(\tau -1| Z_i^t)},
\end{equation} where $S(\tau| Z_i^t) = 1$ for $\tau \leq \tau_{i,l}$. And, we have
\begin{equation} \label{eqn:ef_surv_proof_1}
    \begin{split}
   \lambda_{i, k}^*(t) &= P(T^r = t, k_i = k | T^r \geq t, \mathcal{X}_i, Z_i^t) \\
   &= P(T^r = t, k_i = k | T^r \geq t, Z_i^t)\\
   &= \frac{P(T^r = t, k_i = k|Z_i^t)}{P(T^r > t-1|Z_i^t)} \\
   &= \frac{P(T^r = t, k_i = k|Z_i^t)}{S(t-1|Z_i^t)} \\
   &= \frac{F_k(t|Z_i^t) - F_k(t - 1|Z_i^t)}{S(t-1|Z_i^t)}.
    \end{split}
\end{equation} 
Summing up the hazard function across $b$ events, we have
\begin{equation} \label{eqn:ef_surv_proof_2}
    \begin{split}
   \sum_{k = 1}^b\lambda_{i, k}^*(t) &= \sum_{k = 1}^b\frac{F_k(t|Z_i^t) - F_k(t - 1|Z_i^t)}{S(t-1|Z_i^t)} = \frac{S(t-1|Z_i^t) - S(t|Z_i^t)}{S(t-1|Z_i^t)} = 1 - \frac{S(t|Z_i^t)}{S(t-1|Z_i^t)}.
    \end{split}
\end{equation} 
Finally, rearranging Equation \ref{eqn:ef_surv_proof_2} and plugging it into Equation \ref{eqn:ef_surv_proof_0}, we have
\begin{equation} \label{eqn:ef_surv_proof_3}
    \begin{split}
   S(t|Z_i^t) = \prod_{\tau_{i,l} < \tau \leq t} \frac{S(\tau | Z_i^t)}{S(\tau -1| Z_i^t)} = \prod_{\tau_{i,l} < \tau \leq t} \bigg(1 - \sum_{k = 1}^b \lambda_{i, k}^*(\tau)\bigg).
    \end{split}
\end{equation} 
\end{proof}


\section{Model implementation}
\label{appendix:model_detail}

\subsection{Full set of features}
For MIMIC-III, we utilized a total of 42 features including 2 static features (age and gender) as well as top 40 most frequent time-varying vital signs and laboratory test results (heart rate, respiratory rate, systolic blood pressure, diastolic blood pressure, mean blood pressure, oxygen saturation, temperature, glucose, central venous pressure, hematocrit, potassium, sodium, pulmonary artery pressure systolic, PH, hemoglobin, chloride, CO2 (ETCO2, PCO2, etc.), partial pressure of carbon dioxide, creatinine, blood urea nitrogen, bicarbonate, platelets, anion gap, white blood cell count, magnesium, positive end-expiratory pressure set, calcium, tidal volume observed, partial thromboplastin time, red blood cell count, mean corpuscular volume, prothrombin time inr, prothrombin time pt, fraction inspired oxygen set, peak inspiratory pressure, calcium ionized, phosphate, respiratory rate set, phosphorous, tidal volume set). 

For the in-house data, we utilized a total of 62 features including 11 static features : age, sex, late stage/metastasis indicator, lung-related cancer, breast cancer, pancreatic cancer, thyroid cancer, other Gastrointestinal (GI) cancer, Cancer of Unknown Primary (CUP), Genitourinary (GU), and Gynecologic (gyn) and 51 time-varying features. The set of time-varying features include Body Mass Index (BMI) and various lab tests : PLCO2, PCL, BASO, HCT, MVP, CL, ANEU, EOSP, TP, PGLU, LDH, PLT, MONO, EOS, ALK, ALB, RDW, NEUT, ABASO, SGPT, CRE, ANION, LYMPH, MON, GFR, MCH, LYMP, MCHC, BUN, ALKP, HGB, BASOS, GLOB, K, PNA, AEOSN, MG, TBILI, CA, PK, AEOS, GLU, RBC, NA, CO2, AMONS, PBUN, WBC, MCV, SGOT.

\subsection{Hyperparameter Tuning}  
We used the validation set for optimizing hyperparameters and utilizing early-stopping to avoid potential over-fitting. A random search \citep{Bergstra2012} was utilized for each deep learning based model on the following set of hyperparameters :

\paragraph{SurvLatent ODE (Proposed model)}
Latent trajectory ($Z^t$) dimension : [32, 36, 40, 50],
Input embedding dimension : [40, 50, 60, 70],
Number of layers in the encoder ODE function, $f_{\gamma}(\cdot)$ : [3, 5, 7],
Number of layers in the decoder ODE function, $g_{\phi}(\cdot)$ : [3, 5, 7],
Hidden units in $f_{\gamma}(\cdot)$ and $g_{\phi}(\cdot)$ : [30, 50, 70],
Hidden units in GRU : [30, 50, 70],
Survival loss scale : [50, 100, 150], 
Mini-batch size : [50, 75, 100],
Learning rate : [1e-2],
Hidden units in a cause-specific decoder module : [5, 10, 15],
Number of layers in a cause-specific decoder module : [2,3]

Note that in training, we use \verb|Survival loss scale| to put more focus on the log survival likelihood (i.e. $\text{log}(L_{\text{surv}})$) than ELBO which incorporates data reconstruction loss as well as regularizes initial latent distribution. 

\paragraph{Surv VAE-RNN}
Latent trajectory ($Z^t$) dimension : [32, 36, 40, 50],
Input embedding dimension : [40, 50, 60, 70],
Hidden units in GRU : [30, 50, 70],
Survival loss scale : [50, 100, 150], 
Mini-batch size : [50, 75, 100],
Learning rate : [1e-2],
Hidden units (Cause-specific decoder module) : [5, 10, 15],
Number of layers (Cause-specific decoder module) : [2,3]

\paragraph{Dynamic-Deephit \citep{Lee2020DynamicDeepHit}}
Mini-batch size : [32, 64, 128],
Dropout : [0.4, 0.6, 0.8],
Learning rate : [1e-4, 1e-3, 1e-5],
Hidden units (RNN) : [50, 100, 200, 300],
Hidden units (Fully connected layer) : [50, 100, 200, 300],
Number of layers (RNN) : [2, 4],
Number of layers (Attention) : [2],
Number of layers (Cause-specific module) : [1, 2, 3, 5],
RNN type : [LSTM, GRU],
Activation function (Fully connected layer) : [ReLU, Tanh, ELU],
Activation function (RNN) : [ReLU, Tanh, ELU],
$\beta$ : [0.1, 0.5, 1]

\paragraph{RDSM \citep{Chirag2021DeepParam}}
Number of mixtures : [3, 4, 6, 8],
Event time distribution : [LogNormal, Weibull],
Learning rate : [1e-4, 1e-3],
Hidden units (RNN) : [50, 100, 200],
Number of layers (RNN) : [1, 2, 3, 5],
RNN type: [LSTM, GRU, RNN]
\section{Evaluation metrics}
\label{appendix:eval_metrics}
\subsection{Time-dependent cumulative/dynamic AUC}
The AUC for event $k$ at time $t$ is estimated as follows
\begin{align}
    \label{eqn:auc}
    \widehat{\text{AUC}}_k(t) = \frac{\sum_{(i,j) \in \mathcal{C}(t)}w_i \mathbbm{1}(\hat{F}_k(t|X_j) \leq \hat{F}_k(t|X_i)) }{\sum_{(i,j) \in \mathcal{C}(t)}w_i},
\end{align} where $\mathcal{C}(t)$ is a set of all comparable pairs at time $t$ in the test cohort (i.e. $\mathcal{C}(t) = \{(i,j)~|~\mathbbm{1}(t_i^r \leq t, t_j^r > t, k_i = k) \}$ where $t_i^r$ corresponds to a remaining time to event for patient $i$ who experiences event $k$ before $t$, $t_j^r$ correspond to a remaining time to event for patient $j$ free of any events at $t$ in a competing risks setting, and $w_i$ are inverse probability of censoring weights for patient $i$, non-parametrically estimated by the Kaplan-Meier estimator. Under the independent censoring assumption, Equation (\ref{eqn:auc}) provides consistent estimation of the AUC \citep{Uno2007, Blanche2013, Lambert2016}. We utilized \verb|sksurv| Python package implementation of time-dependent AUC \citep{sksurv}.

\subsection{Time-dependent Brier Score}
We extended the proposed formula in \cite{Graf1999} to measure the time-dependent Brier score for event $k$ in a competing risks setting as follows
\begin{align}
    \text{BS}_k(t) = \frac{1}{N_t}\sum_{i = 1}^{N_t} \mathbbm{1}(t_i^r \leq t, k_i = k)(1 - \hat{F}_k(t|\mathcal{X}_i))^2 w_i + \mathbbm{1}(t_i^r > t)(0 - \hat{F}_k(t|\mathcal{X}_i))^2w_i,
\end{align} where $N_t$ is the number of patients in the held-out test set. We utilized \verb|sksurv| Python package implementation of the time-dependent Brier score \citep{sksurv}.

\section{Dana-Farber Cancer Institute (DFCI) data : cohort summary and additional performance results}
\label{appendix:dvt_detail}

\subsection{Cohort summary} 
\begin{table}[!htb]
\centering
\resizebox{0.4\textwidth}{!}{%
\begin{tabular}{@{}ccccc@{}}
\toprule
\multicolumn{2}{c}{}                      & Train  & Valid  & Test   \\ \midrule
\multicolumn{2}{c}{n}                     & 4797   & 1307   & 2630   \\
\multicolumn{2}{c}{Sex (female)}          & 0.537  & 0.526  & 0.554  \\
\multicolumn{2}{c}{Mean age}              & 60.467 & 60.775 & 60.574 \\
\multicolumn{2}{c}{Event rate (VTE)}      & 0.121  & 0.142  & 0.115  \\
\multicolumn{2}{c}{Event rate (all-cause mortality)}                                                      & 0.463 & 0.481 & 0.464 \\
\multicolumn{2}{c}{Late stage/metastasis} & 0.534  & 0.531  & 0.519  \\ \midrule
\multirow{8}{*}{\begin{tabular}[c]{@{}c@{}}Primary\\ cancer\\ sites\end{tabular}} & Gastrointestinal (GI) & 0.241 & 0.244 & 0.239 \\
            & Lung                        & 0.167  & 0.179  & 0.164  \\
            & Gynecologic (Gyn)           & 0.097  & 0.093  & 0.097  \\
            & Breast                      & 0.088  & 0.083  & 0.088  \\
            & Genitourinary (GU)          & 0.085  & 0.08   & 0.082  \\
            & Unknown                     & 0.035  & 0.034  & 0.037  \\
            & Thyroid                     & 0.025  & 0.026  & 0.02   \\
            & Others                      & 0.263  & 0.26   & 0.273  \\ \bottomrule
\end{tabular}%
}
\caption{The DFCI data cohort summary in terms of number of patients, sex, age, event rates for VTE and all-cause mortality, late stage indicator, and diagnosed cancer types across train, validation, and test sets. Note that sex, late stage indicator, and diagnosed cancer types are shown in proportion.}
\label{tab:dvt_cohort_summary}
\end{table}

\begin{table}[!htb]
\centering
\resizebox{0.5\textwidth}{!}{%
\begin{tabular}{@{}ccc@{}}
\toprule
Cause of death                                & n    & proportion \\ \Xhline{3\arrayrulewidth}
Neoplasms                                     & 3624 & 0.890      \\ \hline
Unknown                                       & 145  & 0.036      \\ \hline
Diseases of the circulatory system            & 85   & 0.021      \\ \hline
\begin{tabular}[c]{@{}c@{}}Diseases of the blood and blood-forming organs and \\ certain disorders involving the immune mechanism\end{tabular} & 56 & 0.014 \\ \hline
Diseases of the respiratory system            & 39   & 0.010      \\ \hline
Certain infectious and parasitic diseases     & 26   & 0.006      \\ \hline
Diseases of the digestive system              & 16   & 0.004      \\ \hline
External causes of morbidity and mortality    & 13   & 0.003      \\ \hline
Endocrine, nutritional and metabolic diseases & 12   & 0.003      \\ \hline
Diseases of the nervous system                & 11   & 0.003      \\ \hline
Mental and behavioural disorders              & 9    & 0.002      \\ \hline
Diseases of the genitourinary system          & 8    & 0.002      \\ \hline
External causes of morbidity and mortality    & 7    & 0.002      \\ \hline
External causes of morbidity and mortality    & 6    & 0.001      \\ \hline
Certain infectious and parasitic diseases     & 4    & 0.001      \\ \hline
Diseases of the musculoskeletal system and connective tissue                                                                                   & 3  & 0.001 \\ \hline
\begin{tabular}[c]{@{}c@{}}Symptoms, signs and abnormal clinical and laboratory findings, \\ not elsewhere classified\end{tabular}             & 3  & 0.001 \\ \hline
External causes of morbidity and mortality    & 3    & 0.001      \\ \Xhline{3\arrayrulewidth}
\end{tabular}%
}
\caption{Summary table for causes of death among those who died (n = 4,070) in the in-house data. We obtained the death causes through National Death Index (NDI) data. We considered all-cause mortality as a competing event for time to VTE event prediction.}
\label{tab:cause_of_death_in_data}
\end{table}

Shown in Table \ref{tab:dvt_cohort_summary} is the VTE cohort summary in terms of number of samples, sex, age, event rates, late stage indicator, and diagnosed cancer types across train, validation, and test sets. And, shown in Table \ref{tab:cause_of_death_in_data} is the summary of death causes obtained through National Death Index (NDI). We considered all-cause mortality as a competing event for time to VTE event prediction. 

\newpage
\subsection{Additional performance results}

\begin{figure*}[!htb]
  \centering
  \includegraphics[width=0.8\textwidth]{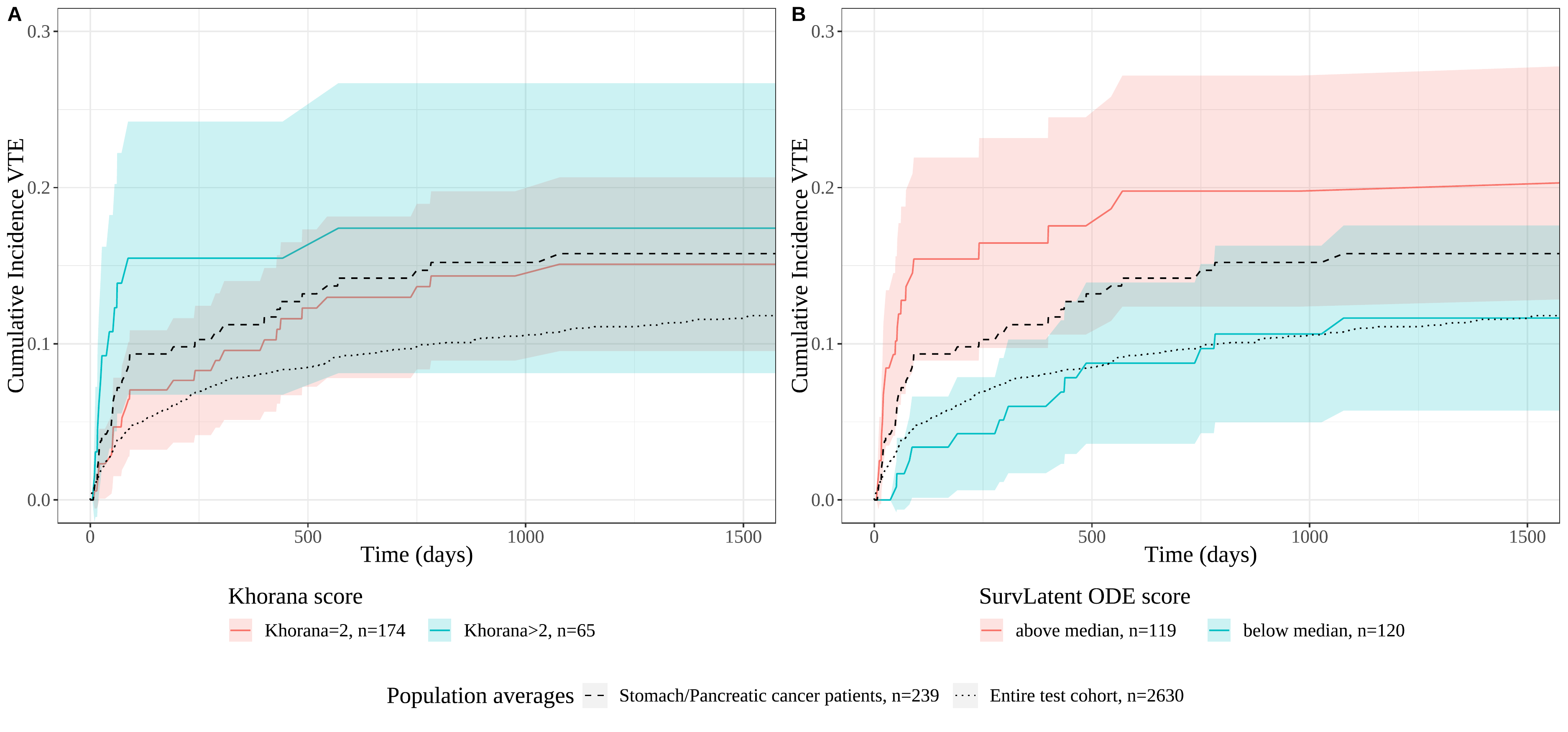}

\caption{\label{fig:khorana_subcancer} \small \textbf{Left} : Khorana-based stratification for patients diagnosed with pancreatic and stomach cancers (n = 239) among the held-out test set. By definition, these patients have at least a Khorana score of 2 due to their high-risk cancer types \citep{Khorana2008}. Cumulative incidence trajectories were estimated by Aalen–Johansen estimator \citep{Johansen1978AnET}, a non-parametric estimator for cumulative incidence, taking into account competing risks. \textbf{Right} : the same cohort of patients was divided into two groups (i.e. above median and below median) based on their cause-specific restricted mean failure time (RMFT) (i.e. $\int_0^{365} \hat{F}_k(t|\mathcal{X}_i)dt$) estimated by SurvLatent ODE. SurvLatent ODE based score was able to capture heterogeneity within the elevated risk group for VTE.}
\end{figure*}

SurvLatent ODE outperformed the Khorana score for the cohort of patients with pancreatic and stomach cancer, where VTE risks are particularly high. SurvLatent ODE identified a low risk subpopulation that had event rates comparable to the population baseline (shown as a dotted line in Fig. \ref{fig:khorana_subcancer}), and substantially lower than the corresponding lowest Khorana score. In this set-up, to stratify risk groups at the prediction time, we computed the cause-specific restricted mean failure time (RMFT) from Day 0 (prediction date) to Day 365 (i.e. $\int_0^{365} \hat{F}_k(t|\mathcal{X}_i)dt$), which is interpreted as the expected number of days lost due to cause $k$ before Day 365 \citep{Mozumder2021}.

\begin{table}[H]
\centering
\resizebox{\textwidth}{!}{%
\begin{tabular}{@{}llllllll@{}}
\toprule
 &
  \begin{tabular}[c]{@{}l@{}}Late stage/metastasis\\ (n = 1366)\end{tabular} &
  \begin{tabular}[c]{@{}l@{}}Lung\\ (n = 432)\end{tabular} &
  \begin{tabular}[c]{@{}l@{}}Gastrointestinal (GI)\\ (n = 389)\end{tabular} &
  \begin{tabular}[c]{@{}l@{}}Stomach/Pancreatic\\ (n = 239)\end{tabular} &
  \begin{tabular}[c]{@{}l@{}}Gynecologic (GYN)\\ (n = 254)\end{tabular} &
  \begin{tabular}[c]{@{}l@{}}Breast\\ (n = 232)\end{tabular} &
  \begin{tabular}[c]{@{}l@{}}Genitourinary (GU)\\ (n = 216)\end{tabular} \\ \midrule
\textbf{\begin{tabular}[c]{@{}l@{}}SurvLatent ODE\\ (Proposed model)\end{tabular}} &
  \textbf{0.733 (0.027)} &
  \textbf{0.765 (0.038)} &
  0.630 (0.067) &
  \textbf{0.793 (0.037)} &
  \textbf{0.580 (0.129)} &
  \textbf{0.914 (0.055)} &
  \textbf{0.874 (0.039)} \\
\begin{tabular}[c]{@{}l@{}}Khorana scores\\ (Imputed)\end{tabular} &
  0.634 (0.031)$^{**}$ &
  0.615 (0.051)$^{*}$ &
  \textbf{0.632 (0.055)} &
  0.632 (0.070)$^{*}$ &
  0.512 (0.131) &
  0.688 (0.100)$^{*}$ &
  0.554 (0.096)$^{**}$ \\ \bottomrule
\end{tabular}%
}
\caption{Performance summary of SurvLatent ODE and Khorana scores on various groups of patients defined by diagnosed cancer types and cancer stage for predicting VTE risk. As a performance metric, we utilized the mean of time-dependent AUC from 25th percentile to 75th percentile of VTE event times. Although model performances vary widely across the groups, SurvLatent ODE outperforms imputed Khorana scores for all the groups except GI group. Standard errors, shown in parenthesis, as well as statistical significance ($* : p < 0.05, ** : p < 0.001$) were obtained exactly in the same manner as in the MIMIC-III experiment (see Table \ref{tab:mimic_perf} caption).}
\label{tab:cancer_type_spec_perf}
\end{table}

\subsection{Latent state explanation}  
\begin{figure*}[!htbp]
  \centering
  \includegraphics[width=0.5\textwidth]{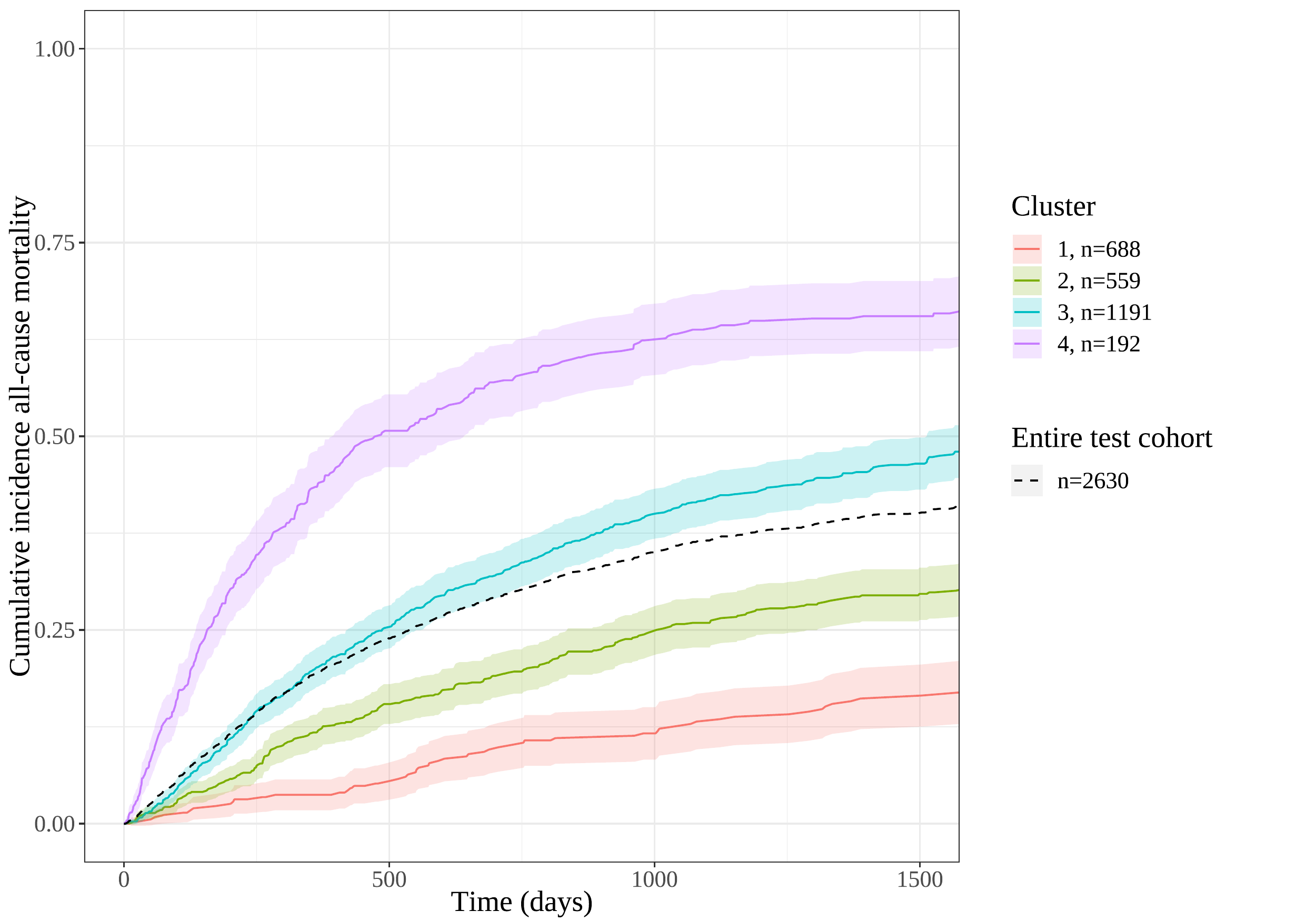}

\caption{\label{fig:death_based_clusters} \small Cumulative Incidence trajectories, estimated by Aalen-Johansen Estimator for the all-cause mortality event across the clusters. The clusters were obtained, based on latent states output from the cause-specific decoder for all-cause mortality.}
\end{figure*}

To obtain latent clusters, we utilized the temporal latent states provided from the cause-specific decoder for VTE just before the fully-connected layer (see Fig. \ref{fig:model_architecture}) at the inference time for the held-out test cohort (n = 2,630). Then, we summed up the latent state of each patient across one year from their latest measurement (i.e. prediction time) and obtained the latent state matrix with dimension of $R^{\text{N}_{\text{test}} \times L}$, where $L$ is the dimension of the latent trajectory. We then ran k-means clustering on the latent state using \verb|sklearn| Python package \citep{sklearn2011}. Finally, we utilized the Aalen–Johansen estimator \citep{Johansen1978AnET} to non-parametrically estimate the survival trajectory of patients in each cluster. We used VTE-based latent clusters to obtain cumulative incidence trajectories for Fig. \ref{fig:latent_clustering_interpretation} and all-cause mortality based latent clusters (i.e. clusters based on the latent states provided from the cause-specific decoder for all-cause mortality) to obtain cumulative incidence trajectories for Fig. \ref{fig:death_based_clusters}.

To interpret each cluster, we first obtained the baseline data for the test cohort at the prediction time using a forward-fill imputation method. Then, we used the standardized baseline data to run logistic regression \citep{statsmodel2010} on each feature conditioning on Khorana scores and see how significantly the feature is associated with each cluster encoded as the binary label (i.e. one-versus-rest). Finally, upon investigating the significance of coefficients for each feature, we dropped a set of features without nominal significance ($p < 0.05$) for any of the clusters. The result is shown in the main text (Fig. \ref{fig:latent_clustering_interpretation}).

\end{document}